\documentclass{article}

\usepackage{microtype}
\usepackage{graphicx}
\usepackage{subfigure}
\usepackage{booktabs} 
\usepackage{multirow}
\usepackage{makecell}
\usepackage[colorlinks,linkcolor=magenta]{hyperref}
\usepackage{hyperref}

\usepackage{hyperref}
\usepackage{bbding}

\usepackage{url}
\usepackage{yfonts}
\usepackage{amssymb}
\usepackage{amsbsy}
\usepackage{amsmath}
\newcommand{\x}{\mathbf{x}}
\newcommand{\y}{\mathbf{y}}
\newcommand{\z}{\mathbf{z}}

\newcommand{\btheta}{\boldsymbol{\theta}}
\newcommand{\bphi}{\boldsymbol{\phi}}
\newcommand{\bpsi}{\boldsymbol{\psi}}
\newcommand{\dkl}{\mathbb{D}_{\rm{KL}}}

\usepackage{booktabs}

\usepackage{color, colortbl, xcolor}
\definecolor{Gray}{gray}{0.9}

\usepackage[accepted]{icml2021}

\icmltitlerunning{A Bit More Bayesian: Domain-Invariant Learning with Uncertainty}

\begin{document}

\twocolumn[
\icmltitle{A Bit More Bayesian: Domain-Invariant Learning with Uncertainty}




\icmlsetsymbol{equal}{*}

\begin{icmlauthorlist}
\icmlauthor{Zehao Xiao}{uva,equal}
\icmlauthor{Jiayi Shen}{uva,equal}
\icmlauthor{Xiantong Zhen}{uva,iiai}
\icmlauthor{Ling Shao}{iiai}
\icmlauthor{Cees G. M. Snoek}{uva}
\end{icmlauthorlist}

\icmlaffiliation{uva}{AIM Lab, University of Amsterdam, The Netherlands}
\icmlaffiliation{iiai}{Inception Institute of Artificial Intelligence, UAE}

\icmlcorrespondingauthor{Z. Xiao}{z.xiao@uva.nl}
\icmlcorrespondingauthor{X. Zhen}{x.zhen@uva.l}
\icmlcorrespondingauthor{C. Snoek}{C.G.M.Snoek@uva.nl}

\icmlkeywords{Machine Learning, ICML}

\vskip 0.3in
]




\printAffiliationsAndNotice{\icmlEqualContribution} 

\begin{abstract}
Domain generalization is challenging due to the domain shift and the uncertainty caused by the inaccessibility of target domain data. In this paper, we address both challenges with a probabilistic framework based on variational Bayesian inference, by incorporating uncertainty into neural network weights.
We couple domain invariance in a probabilistic formula with the variational Bayesian inference. This enables us to explore domain-invariant learning in a principled way. 
Specifically, we derive domain-invariant representations and classifiers, which are jointly established in a two-layer Bayesian neural network. We empirically demonstrate the effectiveness of our proposal on four widely used cross-domain visual recognition benchmarks. 
Ablation studies validate the synergistic benefits of our Bayesian treatment when jointly learning domain-invariant representations and classifiers for domain generalization. Further, our method consistently delivers state-of-the-art mean accuracy on all benchmarks.
\end{abstract}

\section{Introduction}

Learning to improve the generalization of deep neural networks to data out of their training distribution remains 
a fundamental yet challenging problem for machine learning \cite{wang2018learning,bengio2019meta,krueger2020out}. Domain generalization~\cite{muandet2013domain} aims to train a model on several source domains and have it generalize well to unseen target domains. The main challenge stems from the large shift of distributions between the source and target domains, which is further complicated by the prediction uncertainty~\cite{malinin2018predictive} introduced by the inaccessibility to data from the target domains during training. Established approaches learn domain-invariant features by dedicated loss functions \cite{muandet2013domain,li2018learning} or specific architectures \cite{li2017deeper,d2018domain}. 
The state-of-the-art relies on advanced deep neural network backbones, known to degenerate when the test samples are out of the training data distribution~\cite{nguyen2015deep, ilse2019diva}, due to their poorly calibrated behavior~\cite{guo2017calibration,kristiadi2020being}.

Domain generalization is susceptible to uncertainty as the domain shift from the source to the target domain is unknown \textit{a priori}. 
Hence, uncertainty should be taken into account during domain-invariant learning. As deep neural networks are commonly trained by maximum likelihood estimation, they fail to effectively capture model uncertainty. 
This tends to make the models overconfident in their predictions, especially on out-of-distribution data \cite{daxberger2019bayesian}. 
As a possible solution, approximate Bayesian inference offers a natural framework to represent prediction uncertainty~\citep{kristiadi2020being,mackay1992evidence}. It possesses better generalizability to out-of-distribution examples~\citep{louizos2017multiplicative} and provides an elegant formulation to transfer knowledge across different datasets~\citep{nguyen2018variational}. 
Moreover, the prediction uncertainty can be improved, even when Bayesian approximation is only applied to the last network layer~\citep{kristiadi2020being,atanov2019deep}. 
These properties make it appealing to introduce Bayesian learning into the challenging and, as of yet, unexplored scenario of domain generalization. 

%

In this paper, we address domain generalization under a probabilistic framework\footnote{Source code is publicly available at \url{https://github.com/zzzx1224/A-Bit-More-Bayesian.git}.}. To better explore domain-invariant learning, we introduce weight uncertainty to the model by leveraging variational Bayesian inference. 
To this end, we introduce the principle of domain invariance in a probabilistic formulation and incorporate it into the variational Bayesian inference framework. This enables us to explore domain invariance in a principled way to achieve domain-invariant feature representations and classifiers jointly.
To better handle the domain shifts between seen and unseen domains, we explore our method under the meta-learning framework \citep{du2020learning,balaji2018metareg,li2016learning}. 
The meta-learning setting is utilized to expose the model to domain shift and mimic the generalization process by randomly splitting source domains into several meta-source domains and a meta-target domain in each iteration.
We evaluate our method on four widely-used benchmarks for cross-domain object classification. 
Our ablation studies demonstrate the benefit of domain-invariant learning in a probabilistic framework through its synergy with variational Bayesian inference, as well as the advantage of jointly learning domain-invariant feature extractors and classifiers for domain generalization. 
Our method achieves state-of-the-art mean accuracy on all four benchmarks. 
\section{Methodology}
%



\subsection{Preliminaries}
\label{sec:bayes}

In domain generalization, we have $\mathcal{D}{=}{\left\{ D_i \right\}}^{|\mathcal{D}|}_{i=1}{=}\mathcal{S} \cup \mathcal{T}$ as a set of domains, where $\mathcal{S}$ and $\mathcal{T}$ denote the source and target domains, respectively.
$\mathcal{S}$ and $\mathcal{T}$ do not have any overlap besides sharing the same label space.  
Data from the target domains $\mathcal{T}$ is never seen during training. 
For each domain $D_{i} \in \mathcal{D}$, we define a joint distribution $p(\x, \y)$ on $\mathcal{X} \times \mathcal{Y}$, where $\mathcal{X}$ and $\mathcal{Y}$ denote the input space and output space, respectively.
We aim to learn a model $f: \mathcal{X}{\rightarrow} \mathcal{Y}$ in the source domains $\mathcal{S}$ that generalizes well to the target domains $\mathcal{T}$.

We address domain generalization in a probabilistic framework of Bayesian inference by introducing weight uncertainty into neural networks. 
To be specific, we adopt variational Bayesian inference to learn a neural network that is assumed to be parameterized by weights $\btheta$ with a prior distribution $p(\btheta)$ and the posterior distribution $p(\btheta|\x, \y)$, where $(\x, \y)$ are samples from the source domain $\mathcal{S}$. To learn the model, we seek for a variational distribution $q(\btheta)$ to approximate $p(\btheta | \x, \y)$ by minimizing the Kullback-Leibler divergence $\mathbb{D}_{\rm{KL}} \big[ q(\btheta)||p(\btheta|\x, \y)\big]$ between them.
This amounts to minimizing the following objective:
\begin{equation}
\label{elbo1}
\begin{aligned}
\mathcal{L}_{\rm{Bayes}} 
= -\mathbb{E}_{q(\btheta)}[\log p(\y|\x, \btheta)] + \dkl[q(\btheta)||p(\btheta)],
\end{aligned}
\end{equation}
which is also known as the negative value of the evidence lower bound (ELBO) \citep{blei2017variational}. We learn the model on the source domain $\mathcal{S}$ in the hope that it will generalize well to the target domain $\mathcal{T}$.

\citet{kristiadi2020being} show that applying Bayesian approximation to the last layer of a neural network effectively captures model uncertainty. This is also appealing when dealing with uncertainty in domain generalization since applying Monte Carlo sampling to the full Bayesian neural network can be computationally infeasible. 
In this paper, to obtain a model that generalizes well across domains in an efficient way, we incorporate variational Bayesian approximation into the last two network layers to jointly establish domain-invariant feature representations and classifiers. 
More analyses are provided in the experiments and supplementary.
To achieve better domain invariance, we introduce a domain-invariant principle in a probabilistic form under the Bayesian framework.
We use $\bphi$ and $\bpsi$ to denote the parameters of the feature extractor and classifier. We incorporate the domain-invariant principle into the inference of posteriors over $\bpsi$ and $\bphi$, which results in our two-layer Bayesian network. Next we detail the Bayesian domain-invariant learning of the network and the objective for optimization.

\begin{figure}[t]
    \centering
    \includegraphics[width=.99\linewidth]{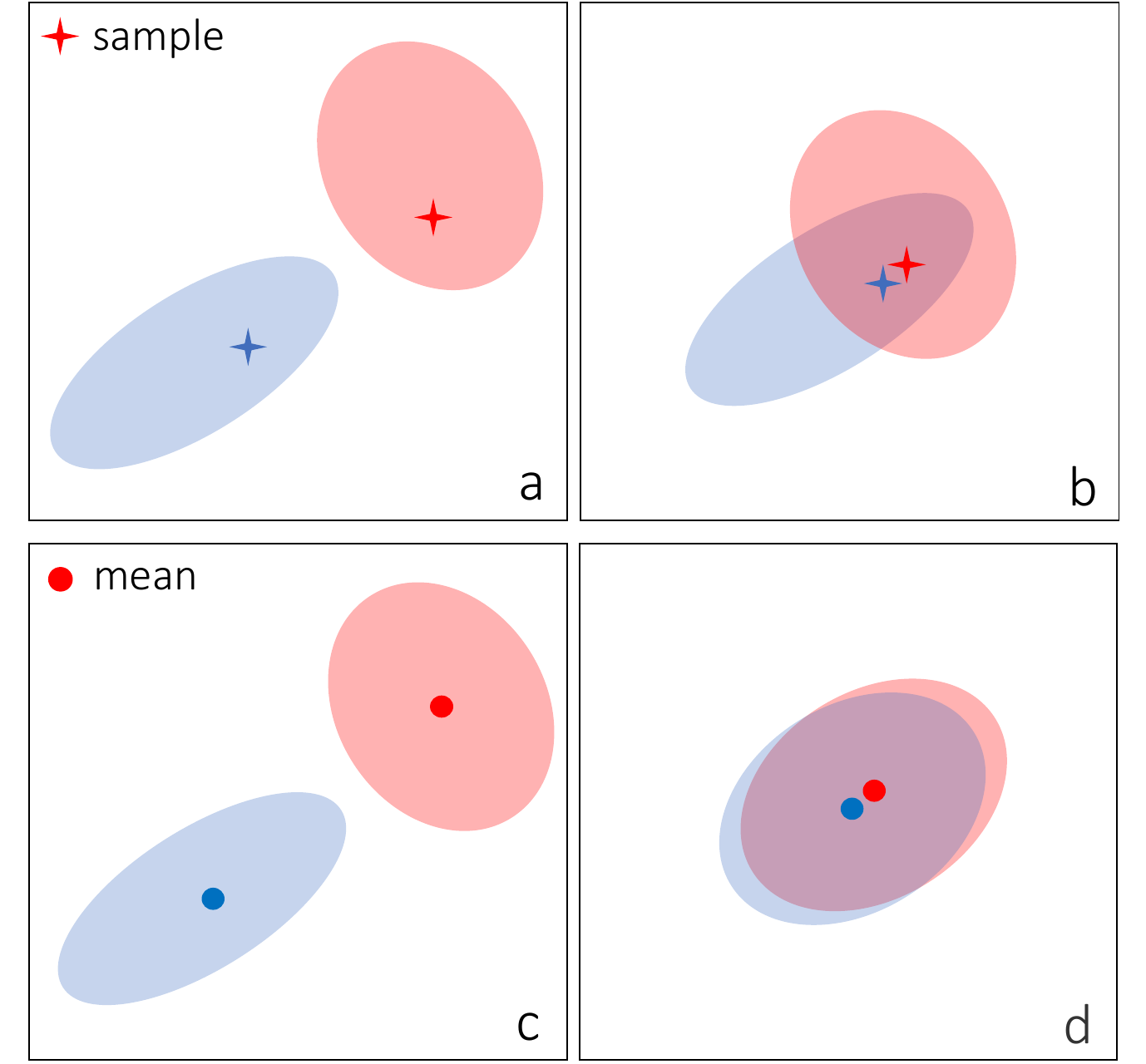}
    \caption{Illustrative contrast between deterministic invariance (top) and probabilistic invariance (bottom), where colors indicate domains. The deterministic invariance tends to minimize the distance between two deterministic samples ($a \rightarrow b$). The samples come from their respective distributions, but with limited distributional awareness. This means that the samples cannot represent the complete distributions. 
    Thus, the deterministic invariance can lead to small distances between samples, but large gaps between distributions, as shown in figure (b). 
    In contrast, the probabilistic invariance directly minimizes the distance between different distributions ($c \rightarrow d$), which yields a better domain invariance as most of the samples in the distributions are taken into account.}
    \label{fig:invariant}
\end{figure}

\subsection{Bayesian Domain-Invariant Learning}
\label{sec:bil}

Existing domain generalizations try to achieve domain invariance by minimizing the distance between intra-class samples from different domains in the hope of reducing the domain gaps \cite{motiian2017unified}. However, individual samples cannot be assumed to be representative of the distributions of samples from a certain domain. 
Therefore, it is preferable to minimize the distributional distance of intra-class samples from different domains, which can directly narrow the domain gap. A contrastive illustration is provided in Figure~\ref{fig:invariant}.
Motivated by this, we propose to address domain generalization under the probabilistic modeling by incorporating weight uncertainty into invariant learning in the variational Bayesian inference framework.

We first introduce the definition of domain invariance in a probabilistic form, which we adopt to learn domain-invariant feature representations and classifiers in a unified way. 
We define a continuous domain space $\mathfrak{D}$ containing a set of  domains $\left\{ D_i \right\}^{|\mathcal{D}|}_{i=1}$ in $\mathcal{D}$ and a domain-transform function $g_{\zeta}(\cdot)$ with parameters $\zeta$ in the domain space. 
$\left\{ D_i \right\}^{|\mathcal{D}|}_{i=1}$ are discrete samples in $\mathfrak{D}$.
Function $g_{\zeta}(\cdot)$ transforms samples $\x$ from a reference domain to a different domain $D_{\zeta}$ with respect to $\zeta$, where $\zeta {\thicksim} q(\zeta)$, and different samples $\zeta$ lead to different sample domains $D_{\zeta} {\thicksim}{\mathfrak{D}}$.

As a concrete example, consider the rotation domains in Rotated MNIST \citep{ghifary2015domain}.
$\mathfrak{D}$ is the rotation domain space, containing images rotated by continuous angles from $0$ to $2\pi$. A sample $\x$ from any domain in $\mathfrak{D}$ can be transferred into a different domain $D_{\zeta}$ by: 
\begin{equation}
g_{\zeta}(\x) =  \left[
 \begin{matrix}
   cos(\zeta) & -sin(\zeta) \\
   sin(\zeta) & -cos(\zeta)
  \end{matrix}
  \right]
  \left[
 \begin{matrix}
   x_{1} \\
   x_{2}
  \end{matrix}
  \right],
\end{equation}
where $\zeta \thicksim$ Uniform$(0, 2\pi)$.
$\zeta$ is the parameter of $g_\zeta(\cdot)$ and different $\zeta$ lead to different rotation angles of the original sample $\x$, which form another domain $D_\zeta$ in $\mathfrak{D}$.
In practice, transformations between domains are more complicated and the exact forms of $g_{\zeta}(\cdot)$ and $q(\zeta)$ are not explicitly known.

Based on the above assumptions about $\mathfrak{D}$, $g_\zeta(\cdot)$ and $q(\zeta)$, we introduce our definition of the probabilistic form of domain invariance as follows.
\newtheorem{define}{\bf Definition}[section]
\begin{define}[Domain Invariance]
\label{def1}
Let $\x_i$ be a given sample from domain $D_i$ in the domain space $\mathfrak{D}$,
and $\x_{\zeta} {=} g_{\zeta}(\x_{i})$ be a transformation of $\x_{i}$ in another domain $D_\zeta$ from the same domain space, where $\zeta {\thicksim} q(\zeta)$. $p_{\btheta}(\y|\x)$ denotes the output distribution of input $\x$ with model $\btheta$.
Model $\btheta$ is domain-invariant in $\mathfrak{D}$ if
\begin{equation}
\label{nai_invariant}
p_{\btheta}(\y_{i}|\x_{i}) = p_{\btheta}(\y_{\zeta}|\x_{\zeta}), \qquad \forall \zeta \thicksim q(\zeta).
\end{equation}
\end{define}
Here, we use $\y$ to represent the output from a neural layer with input $\x$, which can either be the prediction vector from the last layer or the feature vector from the last convolutional layer of a deep neural network.

To adopt the domain-invariant principle in the variational Bayesian inference framework, we reformulate (\ref{nai_invariant}) to an expectation form with respect to $q_{\zeta}$, following \citet{learningprior}:
\begin{equation}
\label{ep_invariant}
p_{\btheta}(\y_{i}|\x_{i}) = \mathbb{E}_{q_{\zeta}} [p_{\btheta}(\y_{\zeta}|\x_{\zeta})].
\end{equation}

According to Definition~\ref{def1}, we use the Kullback-Leibler divergence between the two terms in (\ref{ep_invariant}), $\dkl \big[ p_{\btheta}(\y_{i}|\x_{i})|| \mathbb{E}_{q_\zeta} [p_{\btheta}(\y_{\zeta}|\x_{\zeta})] \big]$, to quantify the domain invariance of the model, which will be zero when the model is domain invariant in the domain space $\mathfrak{D}$. 
To further facilitate the computation, we derive the upper bound of the KL divergence:
\begin{equation}
\label{kl_inequal}
\begin{aligned}
\dkl & \big[ p_{\btheta}(\y_{i}|\x_{i})|| \mathbb{E}_{q_{\zeta}} [p_{\btheta}(\y_{\zeta}|\x_{\zeta})] \big] \\ &\leq \mathbb{E}_{q_{\zeta}} \Big[\dkl \big[ p_{\btheta}(\y_{i}|\x_{i})||p_{\btheta}(\y_{\zeta}|\x_{\zeta}) \big]\Big],
\end{aligned}
\end{equation}
which can be approximated by Monte Carlo sampling as we usually have access to samples from different domains. 
The complete derivation of (\ref{kl_inequal}) is provided in the supplementary material.

We now adopt the probabilistic domain invariance principle in the variational Bayesian approximation of the last two layers with parameters of $\bphi$ and $\bpsi$, respectively. The introduced weight uncertainty results in distributional representations $p_{\bphi}(\z|\x)$ and predictions $p_{\bpsi}(\y|\z)$, based on which we derive domain-invariant learning. 
\paragraph{Invariant Classifier.} For the classifier with parameters $\bpsi$ and input features $\z$, the probabilistic domain-invariant property in (\ref{kl_inequal}) is represented as $\mathbb{E}_{q_{\zeta}} \big[\dkl [ p_{\bpsi}(\y_{i}|\z_{i})||p_{\bpsi}(\y_{\zeta}|\z_{\zeta})]\big]$.
Under the Bayesian framework, the predictive distribution $p_{\bpsi}(\y|\z)$ is obtained by taking the expectation over the distribution of parameter $\btheta$, i.e., $p_{\bpsi}(\y|\z){=}\mathbb{E}_{q(\bpsi)}[p(\y|\z, \bpsi)]$.
As the KL divergence is a convex function \citep{learningprior}, we further extend it to the upper bound:
\begin{equation}
\label{invariant_final}
\begin{aligned}
\mathbb{E}_{q_{\zeta}} & \Big[\dkl \big[ p_{\bpsi}(\y_{i}|\z_{i}) ||p_{\bpsi}(\y_{\zeta}|\z_{\zeta}) \big] \Big] 
\\ & = \mathbb{E}_{q_{\zeta}}\Big[ \dkl \big[ \mathbb{E}_{q(\bpsi)} [p(\y_{i}|\z_{i}, \bpsi)] || \mathbb{E}_{q(\bpsi)} [p(\y_{\zeta}|\z_{\zeta}, \bpsi)] \big] \Big] 
\\ & \leq \mathbb{E}_{q_{\zeta}} \Big[\mathbb{E}_{q(\bpsi)} \dkl \big[ p(\y_{i}|\z_{i}, \bpsi) ||  p(\y_{\zeta}|\z_{\zeta}, \bpsi) \big]\Big], 
\end{aligned}
\end{equation}
which is tractable with unbiased approximation by using Monte Carlo sampling.


In practice, we estimate the expectation over $q(\zeta)$ in an empirical way. Specifically, in each iteration, we choose one domain from the source domains $\mathcal{S}$ as the meta-target domain $D_t$ and the rest are used as the meta-source domains ${\left\{D_s \right\}}^{S}_{s=1}$, where $S{=}|\mathcal{S}|-1$.
Then we use a batch of samples $\x_s$ from each meta-source domain in the same category as $\x_t$ to approximate $\x_\zeta {=} g_{\zeta}(\x_{t})$. 

Thereby, the domain-invariant classifier is established by minimizing: 
\begin{equation}
\begin{aligned}
\label{Li_y}
\frac{1}{SN}\sum_{s=1}^S\sum_{i=1}^N  \mathbb{E}_{q(\bpsi)}\Big[ \dkl \big[ p(\y_{t}|\z_{t}, \bpsi) ||  p(\y_{s}^{i}|\z_{s}^{i}, \bpsi) \big]\Big],
\end{aligned}
\end{equation}
where ${\left\{\z_{s}^{i} \right \}}^{N}_{i=1}$ are representations of samples $\x_s$ from $D_{s}$, which are in the same category as $\x_{t}$. 



The variational Bayesian inference enables us to flexibly incorporate our domain invariant principle into different layers of the neural network. To enhance the domain invariance, we obtain domain-invariant representations by adopting the principle in the penultimate layer.

\paragraph{Invariant Representations.} 
To obtain domain-invariant representations $p(\z|\x)$ with feature extractor $\bphi$, we extend the quantification of our probabilistic domain invariance in (\ref{kl_inequal}) to $\mathbb{E}_{q_{\zeta}} \big[\dkl [ p_{\bphi}(\z_{i}|\x_{i})||p_{\bphi}(\z_{\zeta}|\x_{\zeta})]\big]$.
Based on the Bayesian layer, the probabilistic distribution of features $p_{\bphi}(\z|\x)$ will be a factorized Gaussian distribution if the posterior of $\bphi$ is as well. 
We illustrate this as follows. 
Let $\bphi$ be the last Bayesian layer in the feature extractor with a factorized Gaussian posterior and $\x$ be the input feature of $\bphi$. 
The posterior of the activation $\z$ is also a factorized Gaussian \citep{kingma2015variational}:
\begin{equation}
\begin{aligned}
\label{distributionz}
&q(\phi_{i,j}) \thicksim \mathcal{N}(\mu_{i,j}, \sigma^2_{i,j}) ~~~\forall \phi_{i,j} \in \bphi 
\\&
\Rightarrow 
p(z_{j}|\x, \bphi) \thicksim{\mathcal{N}(\gamma_{j}, \delta_{j}^2)}, \\
&\gamma_{j}=\sum_{i=1}^{N}x_{i}\mu_{i,j}, ~~~~ \text{and} ~~~~\delta_{j}^2=\sum_{i=1}^{N}x_{i}^2\sigma_{i,j}^2,
\end{aligned}
\end{equation}
where $z_{j}$ denotes the $j$-th element in $\z$, likewise for $x_{i}$, and $\phi_{i,j}$ denotes the element at position $(i,j)$ in $\bphi$.
Thus, we assume the posterior of the last Bayesian layer in the feature extractor has a factorized Gaussian distribution. 
Then, it is easy to obtain the Bayesian domain invariance of the feature extractor.

In a similar way to (\ref{Li_y}), we estimate the expectation over $q(\zeta)$ empirically. 
Therefore, the domain-invariant representations are established by minimizing:
\begin{equation}
\label{Li_z}
\frac{1}{SN}\sum_{s=1}^S \sum_{i=1}^N \Big[ \dkl \big[ p(\z_{t}|\x_{t}, \bphi) ||  p(\z_{s}^{i}|\x_{s}^{i}, \bphi) \big]\Big],
\end{equation}
where ${\left\{\x_{s}^{i} \right \}}^{N}_{i=1}$ are from $D_{s}$, and denote the samples in the same category as $\x_{t}$. 
More details and an illustration of the Bayesian domain-invariant learning are provided in the supplementary material.

\subsection{Objective Function} 
Having the Bayesian treatment for the last two layers with respect to parameters $\bphi$ and $\bpsi$, the $\mathcal{L}_{\rm{Bayes}}$ in (\ref{elbo1}) is instantiated as the objective with respect to $\bpsi$ and $\bphi$ as follows: 
\begin{equation}
\begin{aligned}
{\mathcal{L}}_{\rm{Bayes}}
& = -\mathbb{E}_{q(\bpsi)}\big[ \mathbb{E}_{q(\bphi)} [\log p(\y|\x, \bpsi, \bphi)] \big]
\\ & + \dkl[q(\bpsi)||p(\bpsi)]  + \dkl[q(\bphi)||p(\bphi)].
\label{bayes}
\end{aligned}
\end{equation}
The detailed derivation of (\ref{bayes}) is provided in the supplementary material.

By integrating (\ref{Li_y}) and (\ref{Li_z}) into (\ref{bayes}), we obtain the Bayesian domain-invariant learning objective as follows:
%
\begin{equation}
\begin{aligned}
    {\mathcal{L}}_{\rm{BIL}} & =  \frac{1}{L}\sum_{\ell=1}^L \Big[ \frac{1}{M}\sum_{m}^M [-\log p(\y_t|\x_t, \bpsi^{(\ell)}, \bphi^{(m)})] \\ &
    + \frac{1}{SN}\sum_{s=1}^S 
    \sum_{i=1}^N 
    \big[\lambda_{\bpsi} \dkl[p(\y_t|\z_t, \bpsi^{(\ell)})||p(\y^i_{s}|\z^i_{s}, \bpsi^{(\ell)})] \\ & 
    + \lambda_{\bphi} 
    \dkl[p(\z_t|\x_t, \bphi)||p(\z^i_{s}|\x^i_{s}, \bphi)] \big] \\&
    + \dkl [q(\bpsi)||p(\bpsi)] + \dkl [q(\bphi)||p(\bphi)] \Big],
\end{aligned}
\end{equation}
where $\lambda_{\bpsi}$ and $\lambda_{\bphi}$ are hyperparameters that control the domain-invariant terms. $\x_t$ and $\z_t$ denote the input and its feature from $D_t$, and $\x^i_s$ and $\z^i_s$ are from $D_s$. The posteriors are set to factorized Gaussian distributions, i.e.,
$q(\bpsi){=}{\mathcal{N}(\boldsymbol{\mu}_{\bpsi}, \boldsymbol{\sigma}_{\bpsi}^{2})}$ and $q(\bphi){=}{\mathcal{N}(\boldsymbol{\mu}_{\bphi}, \boldsymbol{\sigma}_{\bphi}^{2})}$.
We adopt the reparameterization trick to draw Monte Carlo samples~\citep{kingma2013auto} as
$\bpsi^{(\ell)} {=} \boldsymbol{\mu}_{\bpsi} + \epsilon^{(\ell)} * \boldsymbol{\sigma}_{\bpsi}$,
where $\epsilon^{(\ell)}\thicksim{\mathcal{N}(0, I)}$. Likewise, we draw the Monte Carlo samples $\bphi^{(m)}$ for $q(\bphi)$. 

When implementing our Bayesian invariant learning, to increase the flexibility of the prior distribution in our Bayesian layers, we
place a scale mixture of two Gaussian distributions as the priors $p(\bpsi)$ and $p(\bphi)$ \citep{Blundell2015WeightUI}:
\begin{equation}
\label{scalemixture}
    {\pi\mathcal{N}(0, \boldsymbol{\sigma}_{1}^{2})}+{(1-\pi)\mathcal{N}(0, \boldsymbol{\sigma}_{2}^{2})},
\end{equation}
where $\boldsymbol{\sigma}_1$, $\boldsymbol{\sigma}_2$ are set according to \citep{Blundell2015WeightUI} and $\pi$ is chosen by cross-validation. 
More experiments and analyses on the hyperparameters $\lambda_{\bpsi}$, $\lambda_{\bphi}$ and $\pi$ are provided in the supplementary material.


\section{Related Work}
To address the domain shifts, domain adaptation \citep{saenko2010adapting,long2015learning,wang2020dictionary} and domain generalization \citep{muandet2013domain,li2017deeper} are the two main  settings.
Domain adaptation has a key assumption that the data from the target domain is accessible during training, which is often invalid in realistic applications.
By contrast, no target domain data is available during training in domain generalization, which introduces more prediction uncertainty into the problem and makes it more challenging.
Our method is developed specifically for domain generalization.

One solution for domain generalization is to generate more source domain data to increase the probability of covering the data in the target domains \citep{shankar2018generalizing,volpi2018generalizing}. \citet{shankar2018generalizing} augment the data by perturbing the input images with adversarial gradients generated by an auxiliary classifier. 
\citet{qiao2020learning} propose an even more challenging scenario of domain generalization named single domain generalization, which only has one source domain, and introduce an adversarial domain augmentation method that creates ``fictitious'' yet ``challenging'' data. Recently, \citet{zhou2020learning} employ a generator to synthesize data from pseudo-novel domains to augment the source domains, maximizing the distance between the source and pseudo-novel domains as measured by optimal transport \citep{peyre2019computational}. 

Another solution for domain generalization involves learning domain-invariant features \citep{d2018domain,li2018domain,li2017deeper}. 
\citet{muandet2013domain} propose domain-invariant component analysis to learn invariant transformations by minimizing the dissimilarity across domains. 
\citet{louizos2015variational} learn invariant representations by a variational auto-encoder  \citep{kingma2013auto}, introducing Bayesian inference into invariant feature learning.
Both \citet{dou2019domain} and \citet{seo2019learning} achieve a similar goal by introducing two complementary losses and employing multiple normalizations. \citet{Li2019EpisodicTF} propose an episodic training algorithm to obtain both a domain-invariant feature extractor and classifier.
 \citet{zhao2020domain} propose entropy-regularization to learn domain-invariant features. \citet{seo2019learning} and \citet{zhoudomain} design  normalizations to combine different feature statistics and embed the samples into a domain-invariant feature space.

Meta-learning has also been considered for domain generalization. \citet{li2018learning} introduce a gradient-based method, i.e., model agnostic meta-learning~\citep{finn2017model}, for domain generalization. \citet{balaji2018metareg} meta-learn a regularization function, making their model robust to domain shifts. \citet{du2020learning} propose the meta-variational information bottleneck to learn domain-invariant representations through episodic training.

\citet{gulrajani2020search} find that with careful implementation, empirical risk minimization methods outperform many state-of-the-art models in domain generalization. They claim that model selection is non-trivial for domain generalization and algorithms for this task should specify their own model selection criteria, which is important for the completeness and comparability of the method.
In this paper, we implement our method based on the pretrained ResNet-18, as done for our baseline and most methods we compare against in Section~\ref{experiment}.

\citet{motiian2017unified} previously considered representation alignment across domains in the same class. They propose a classification and contrastive semantic alignment loss based on the L2 distance between deterministic features. 
Different from them, we exploit Bayesian neural networks to learn domain-invariant representations by minimizing the distance between probabilistic distributions. 
%

Bayesian neural networks have not yet been explored for domain generalization. 
Our method introduces variational Bayesian approximation to both the feature extractor and classifier of the neural network in conjunction with the newly introduced domain-invariant principle for domain generalization. The resultant Bayesian domain-invariant learning combines the representational power of deep neural networks and variational Bayesian inference.

\section{Experiments} \label{experiment}

\subsection{Datasets and Settings}
\label{setting}

We conduct our experiments on four widely used benchmarks for domain generalization and report the mean classification accuracy on target domains.

\textbf{PACS}\footnote{\url{https://domaingeneralization.github.io/}}~\citep{li2017deeper} consists of 9,991 images of seven classes from four domains -- \textit{photo}, \textit{art-painting}, \textit{cartoon} and \textit{sketch}. We follow the ``leave-one-out'' protocol from~\citep{li2017deeper,li2018domain,carlucci2019domain}, where the model is trained on any three of the four domains, which we call source domains, and tested on the last (target) domain. 

\textbf{Office-Home}\footnote{\url{https://www.hemanthdv.org/officeHomeDataset.html}} \citep{venkateswara2017deep} also has four domains: \textit{art}, \textit{clipart}, \textit{product} and \textit{real-world}. There are about 15,500 images of 65 categories for object recognition in office and home environments. We use the same experimental protocol as for PACS.

\textbf{Rotated MNIST}\footnote{\url{http://yann.lecun.com/exdb/mnist/}} and \textbf{Fashion-MNIST}\footnote{\url{https://www.kaggle.com/zalando-research/fashionmnist}} are introduced in \citep{piratla2020efficient} for evaluating domain generalization.
For fair comparison, we follow their recommended settings and randomly select a subset of 2,000 images from MNIST and 10,000 images from Fashion-MNIST, which are considered to have been rotated by $0^\circ$. The subset of images is then rotated by $15^\circ$ through $75^\circ$ in intervals of $15^\circ$, creating five source domains. The target domains are created by rotations of $0^\circ$ and $90^\circ$. 
These datasets allow us to demonstrate the generalizability of our model by comparing its performance on in-distribution and out-of-distribution data.

\begin{table*}[t]
\caption{Ablation study on PACS. The ``$\checkmark$'' and ``$\times$'' in the ``Bayesian'' column indicate whether the classifier $\bpsi$ and feature extractor $\bphi$ are Bayesian layers or deterministic layers. 
In the ``Invariant'' column, they indicate whether the domain-invariant learning is introduced into the classifier and the feature extractor.
Note that with ``$\checkmark$'' in the Bayesian column, the invariant column denotes Bayesian domain-invariant learning. Otherwise it denotes the deterministic one.
The results show that both Bayesian and domain-invariant learning benefit domain generalization, but our Bayesian domain-invariant learning is better.
We obtain the best performance with Bayesian domain-invariant learning in both the classifier and feature extractor. 
We also provide the visualizations of the features for each of the settings in Figure~\ref{fig:visual}.
} 
\vspace{2mm}
\centering
\label{ablation}
		\setlength\tabcolsep{4pt} 
\begin{tabular}{cccccccccc}
\toprule
 & \multicolumn{2}{c}{\textbf{Classifier $\bpsi$}} & \multicolumn{2}{c}{\textbf{Feature extractor $\bphi$}} & \multicolumn{5}{c}{\textbf{PACS}}\\
\cmidrule(lr){2-3} \cmidrule(lr){4-5} \cmidrule(lr){6-10}
ID &
  Bayesian &
  Invariant &
  Bayesian &
  Invariant &
  Photo &
  Art-painting &
  Cartoon &
  Sketch &
  \textit{Mean} \\ 
  \midrule
(a) & $\times$     & $\times$     & $\times$     & $\times$  & 92.85 {\scriptsize $\pm$0.21}   & 75.12 {\scriptsize $\pm$0.48} & 77.44  {\scriptsize $\pm$0.26} & 75.72 {\scriptsize $\pm$0.47} & 80.28 {\scriptsize $\pm$0.42} \\
\hline
(b) & $\checkmark$ & $\times$     & $\times$     & $\times$  & 93.89 {\scriptsize $\pm$0.29} & 77.88 {\scriptsize $\pm$0.53} & 78.20 {\scriptsize $\pm$0.39} & 77.75 {\scriptsize $\pm$0.75} & 81.93 {\scriptsize $\pm$0.22}\\
 (c) &
  $\times$ &
  $\checkmark$ &
  $\times$ &
  $\times$ & 93.95 {\scriptsize $\pm$0.51} & 80.03 {\scriptsize $\pm$0.72}  & 78.03 {\scriptsize $\pm$0.77} & 77.83 {\scriptsize $\pm$0.52} & 82.46 {\scriptsize $\pm$0.67}
   \\
 (d) &
  $\checkmark$ &
  $\checkmark$ &
  $\times$ &
  $\times$ &
  95.21 {\scriptsize $\pm$0.26}  & 81.25 {\scriptsize $\pm$0.76} & 80.67 {\scriptsize $\pm$0.73} & 79.31 {\scriptsize $\pm$0.94} & 84.11 {\scriptsize $\pm$0.39} \\ 
  \hline
(e) & $\times$     & $\times$ & $\checkmark$ & $\times$ & 92.81 {\scriptsize $\pm$0.35} &
  78.66 {\scriptsize $\pm$0.56} &
  77.90 {\scriptsize $\pm$0.40} &
  78.72 {\scriptsize $\pm$0.86} &
  82.02 {\scriptsize $\pm$0.26}  \\
 (f) & $\times$     & $\times$ & $\times$     & $\checkmark$   & 94.17 {\scriptsize $\pm$0.35} & 79.75 {\scriptsize $\pm$0.68}  & 79.51 {\scriptsize $\pm$0.98} & 78.31 {\scriptsize $\pm$1.11} & 82.94 {\scriptsize $\pm$0.53} \\
 (g) & $\times$     & $\times$ & $\checkmark$     & $\checkmark$ & 95.15 {\scriptsize $\pm$0.26} & 80.96 {\scriptsize $\pm$0.69} & 79.57 {\scriptsize $\pm$0.85} & 79.15 {\scriptsize $\pm$0.98} & 83.71 {\scriptsize $\pm$0.65}\\ \hline
(h) & $\checkmark$ & $\times$ & $\checkmark$     & $\times$  & 93.83 {\scriptsize $\pm$0.19} & 82.13 {\scriptsize $\pm$0.41} & 79.18 {\scriptsize $\pm$0.48} & 79.03 {\scriptsize $\pm$0.78} & 83.54 {\scriptsize $\pm$0.34} \\
 (i) & $\times$     & $\checkmark$     & $\times$ & $\checkmark$ & 94.12 {\scriptsize $\pm$0.22} & 80.52 {\scriptsize $\pm$0.61} & 80.39 {\scriptsize $\pm$0.81} & 78.53 {\scriptsize $\pm$0.95} & 83.39 {\scriptsize $\pm$0.52} \\
 (j) &
  $\checkmark$ &
  $\checkmark$ &
  $\checkmark$ &
  $\checkmark$ &
  \textbf{95.97} {\scriptsize $\pm$0.24}&
  \textbf{83.92} {\scriptsize $\pm$0.71}&
  \textbf{81.61} {\scriptsize $\pm$0.59}&
  \textbf{80.31} {\scriptsize $\pm$0.91}&
  \textbf{85.45} {\scriptsize $\pm$0.24}\\ 
  \bottomrule
\end{tabular}
\vspace{-1mm}
\end{table*}

\begin{figure*}[h]
\centering
\includegraphics[width=\linewidth]{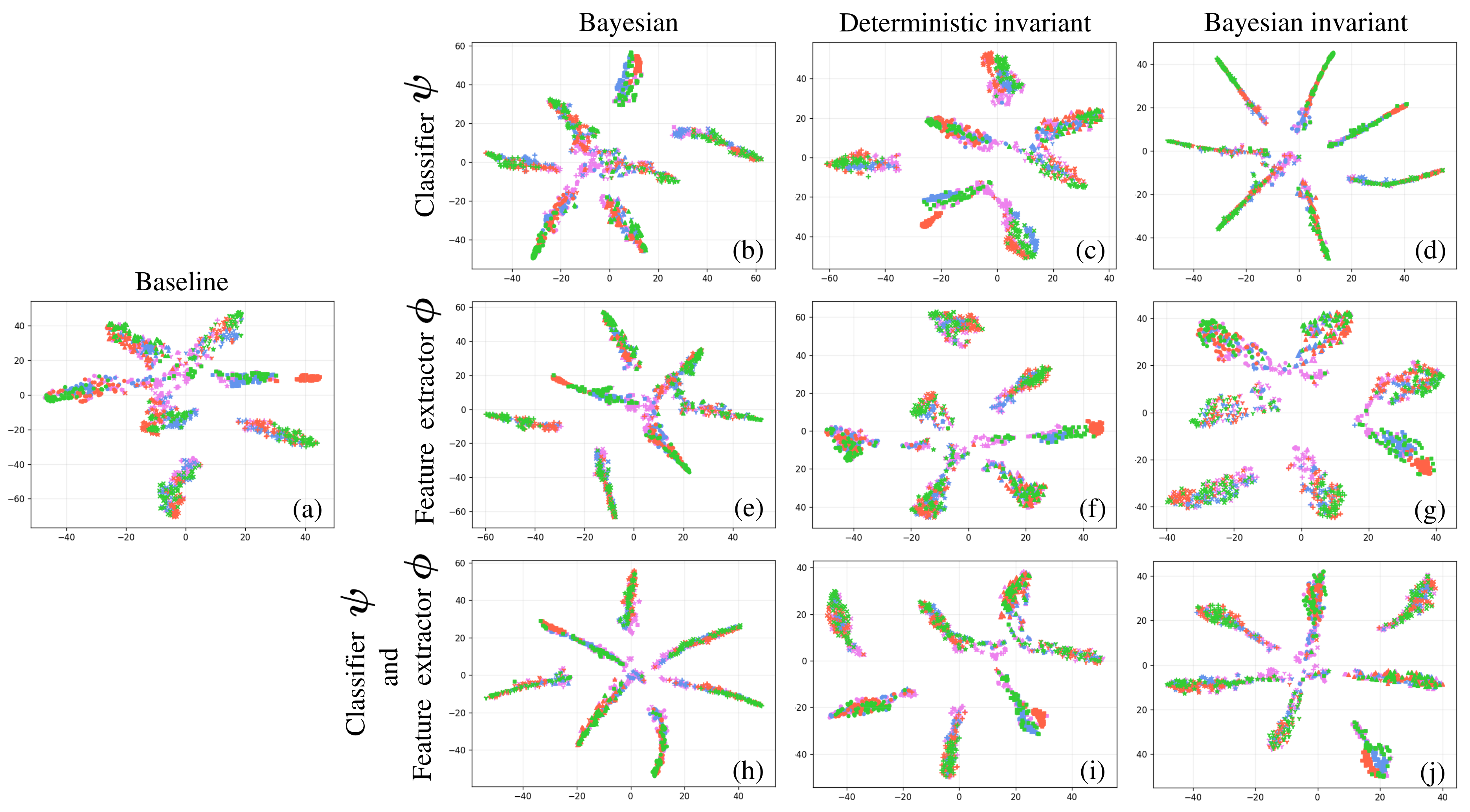}
\caption{Illustration of the benefit of Bayesian domain-invariant learning by visualization of the feature representations. Colors denote domains, where the target domain ``art-painting'' is in violet, and shapes indicate classes. Our Bayesian domain-invariant learning (shown in the right column), especially when employed in both the classifier and feature extractor ((j)), achieves better results than the other cases shown in the ten subfigures, which correspond to the ten settings in Table \ref{ablation} (identified by ID). 
}
\label{fig:visual}
\vspace{-2mm}
\end{figure*}

\textbf{Settings.} For all four benchmarks, we employ a ResNet-18 \citep{he2016deep} pretrained on ImageNet \citep{deng2009imagenet} as the backbone.
During training we use Adam optimization \citep{kingma2014adam} with a learning rate of 0.0001, and train for 10,000 iterations.
In each iteration we choose one source domain as the meta-target domain. 
The batch size is 128. 
To fit the memory footprint, we choose a maximum number of samples per category and target domain to implement the domain-invariant learning, i.e., sixteen for PACS, Rotated MNIST and Fashion-MNIST, and four for Office-Home. 
We select $\lambda_{\bphi}$ and $\lambda_{\bpsi}$ based on validation set performance and summarize their influence in the supplementary material. 
The optimal values of $\lambda_{\bphi}$ and $\lambda_{\bpsi}$ are 0.1 and 100.
Parameters $\sigma_1$ and $\sigma_2$ in (\ref{scalemixture}) are set to 0.1 and 1.5. The model with the highest validation set accuracy is employed for evaluation on the target domain. 

\subsection{Results}
\label{section:sec_abl}

We first conduct an ablation study on PACS to investigate the effectiveness of our Bayesian invariant layer for domain generalization. Since the major contributions of this work are the Bayesian treatment and the probabilistic domain-invariant principle, we evaluate their effect by individually incorporating them into the classifier -- the last layer -- $\bpsi$ and the feature extractor -- the penultimate layer -- $\bphi$. 
The results are shown in Table \ref{ablation}, with a corresponding t-SNE visualization of the features learned by various settings in Figure~\ref{fig:visual}, following \citet{du2020learning}.
We also provide a comparison with the state-of-the-art methods on four widely used benchmarks.
More results and visualizations are provided in the supplementary material.

\textbf{Benefits of Bayesian Invariant Classifier.}
In Table~\ref{ablation}, rows (a) to (d) demonstrate the benefits of the Bayesian invariant classifier. 
Row (a) serves as our baseline model, which is a vanilla deep convolutional network without any Bayesian treatment or domain-invariant loss. 
The backbone is also a ResNet-18 pretrained on ImageNet. 
Rows (b), (c) and (d) show the performance with a Bayesian classifier, a deterministic domain-invariant classifier and our Bayesian-invariant classifier.
By comparing (b) and (a), it is clear that the Bayesian treatment for the classifier improves the performance, especially in the ``art-painting'' and ``sketch'' domains. The deterministic invariant property, as shown in row (c), also benefits the performance.
Nevertheless, our Bayesian invariant learning based on the probabilistic framework performs better.
The results demonstrate that the Bayesian invariant learning  enhances the robustness of Bayesian neural networks on out-of-distribution data and  better leverages domain-invariance than the deterministic invariant model.

The three subfigures in the first row of Figure~\ref{fig:visual} also demonstrate the benefits of the Bayesian invariant classifier.
The Bayesian treatment enlarges the inter-class distance in all domains, as shown in Figure~\ref{fig:visual} (b).
The Bayesian layer incorporates weight uncertainty into the predictions and improves their diversity, which enhances the classification performance on the source domains as well as the generalization to the target domain. 
The deterministic invariant classifier tends to minimize the distance between samples with the same label. However, the effect is mainly on the source domain, without obvious impact on the target domain (pink samples), as shown in Figure~\ref{fig:visual} (c).
By introducing uncertainty into both the classification and domain-invariant procedure, our Bayesian domain-invariant classifier further enlarges the inter-class distance in all domains and better generalizes the domain-invariant property to the target domain (compare  Figure~\ref{fig:visual} (d) to (b) and (c)).

\textbf{Benefits of Bayesian Invariant Feature Extractor.}
The benefit of our Bayesian domain-invariant principle for the feature extractor is demonstrated in rows (e), (f), and (g) of Table~\ref{ablation}.
Similar to the classifier, introducing Bayesian inference into the feature extractor (comparing row (e) with row (a)) 
brings a good accuracy improvement. 
By comparing (g) to (e) and (f), we observe that our Bayesian invariant layer achieves consistently better accuracy than the deterministic invariant model and regular Bayesian layer.

The subfigures in the second row of Figure~\ref{fig:visual} further demonstrate the effectiveness of the Bayesian domain-invariant feature extractor. 
Similar to the classifier, introducing the Bayesian framework into the feature extractor also enlarges the inter-class distance of all domains by introducing model uncertainty, as shown in (e).
The deterministic invariant feature extractor can also minimize the intra-class distance between different domains.
However, while the effect 
is more obvious than the deterministic domain-invariant classifier, it 
is still poor on the target domain (Figure~\ref{fig:visual} (f)).
Compared to (e) and (f), the Bayesian invariant feature extractor in (g) tends to further maximize the inter-class distance while minimizing the intra-class distance between different domains, which obviously enlarges the inter-class distance between samples from the target domain.
Since the Bayesian invariant feature extractor achieves domain invariance in a probabilistic way, more uncertainties are taken into account, enabling the model to achieve better generalization on the target domain. 
Note that although the Bayesian invariant classifier in (d) and feature extractor in (g) achieve domain-invariant learning from different directions of the feature space, they both improve the performance on the target domain and do not conflict with each other.

\begin{table*}[t]
\caption{Performance of different priors on PACS.
``Standard'' denotes the standard Gaussian prior while ``Mixture'' denotes the scale mixture prior. 
}
\vspace{1mm}
\centering
\label{prior}
		\setlength\tabcolsep{4pt} 
\begin{tabular}{lccccc}
\toprule
Prior    & Photo    & Art-painting         & Cartoon        & Sketch         & \textit{Mean}       \\ 
    \midrule
Standard & 95.17 {\scriptsize $\pm$0.25} & 81.95 {\scriptsize $\pm$0.42} & 79.41 {\scriptsize $\pm$0.82} & 77.88 {\scriptsize $\pm$0.57} & 83.61 {\scriptsize $\pm$0.12} \\
Mixture & 95.97 {\scriptsize $\pm$0.24} & 83.92 {\scriptsize $\pm$0.71} & 81.61 {\scriptsize $\pm$0.59} & 80.31 {\scriptsize $\pm$0.91} & 85.45 {\scriptsize $\pm$0.24}  \\ 
\bottomrule
\end{tabular}
\vspace{-5mm}
\end{table*}

\begin{table*}[t]
\caption{Effect of more Bayesian layers on PACS. 
The ``Bayesian'' and ``Invariant'' columns indicate whether the penultimate layer $\bphi '$ in the feature extractor has a Bayesian and/or domain-invariant property.
More Bayesian layers benefit the performance while excessive domain-invariant learning is harmful. 
}
\vspace{2mm}
\centering
\label{2layers}
		\setlength\tabcolsep{4pt} 
\begin{tabular}{ccccccc}
\toprule
\multicolumn{2}{c}{$\bphi '$} & \multicolumn{5}{c}{\textbf{PACS}}\\
\cmidrule(lr){1-2}
\cmidrule(lr){3-7}
Bayesian & Invariant & Photo                                  & Art-painting                           & Cartoon                       & Sketch                        & \textit{Mean}  \\ \midrule
$\times$          & $\times$           & \cellcolor[HTML]{FFFFFF}\textbf{95.97} {\scriptsize $\pm$0.24} & \cellcolor[HTML]{FFFFFF}\textbf{83.92} {\scriptsize $\pm$0.71} & \cellcolor[HTML]{FFFFFF}81.61 {\scriptsize $\pm$0.59} & \cellcolor[HTML]{FFFFFF}80.31 {\scriptsize $\pm$0.91} & \cellcolor[HTML]{FFFFFF}85.45 {\scriptsize $\pm$0.24} \\
$\checkmark$      & $\times$           & 95.69 {\scriptsize $\pm$0.20} & 83.28 {\scriptsize $\pm$0.74} & \textbf{82.06} {\scriptsize $\pm$0.25} & \textbf{81.00} {\scriptsize $\pm$0.55} & \textbf{85.51} {\scriptsize $\pm$0.13} \\
$\checkmark$& $\checkmark$ & 95.72 {\scriptsize $\pm$0.29} & 82.33 {\scriptsize $\pm$0.47} & 81.10 {\scriptsize $\pm$0.73} & 80.67 {\scriptsize $\pm$1.01} & \cellcolor[HTML]{FFFFFF}84.96 {\scriptsize $\pm$0.34} \\ \bottomrule
\end{tabular}
\vspace{-4mm}
\end{table*}

\begin{table*}[t!]
\caption{Comparison on PACS. Our method achieves the best performance on the ``Cartoon'' domain, is competitive on the other three domains and obtains the best overall mean accuracy.}
\vspace{2mm}
\centering
\label{pacs}
		\setlength\tabcolsep{4pt} 
\begin{tabular}{llllll}
\toprule
    & Photo    & Art-painting         & Cartoon        & Sketch         & \textit{Mean}       \\ 
    \midrule
Baseline   & 92.85    & 75.12         & 77.44          & 75.72          & 80.28         \\
\citet{carlucci2019domain}      & 96.03   & 79.42                 & 75.25          & 71.35          & 80.51         \\
\citet{dou2019domain}       & 94.99   & 80.29                 & 77.17          & 71.69          & 81.04         \\
\citet{zhao2020domain} & \textbf{96.65} & 80.70 & 76.40 & 71.77 & 81.38 \\
\citet{piratla2020efficient}         & 94.10          & 78.90         & 75.80          & 76.70          & 81.40         \\
\citet{chattopadhyay2020learning}       & 93.35          & 76.90          & 80.38          & 75.21          & 81.46         \\
\citet{Li2019EpisodicTF}    &93.90      & 82.10               & 77.00          & 73.00          & 81.50         \\
\citet{balaji2018metareg}    & 95.50    & 83.70                & 77.20          & 70.30          & 81.68         \\
\citet{zhou2020learning}    & 96.20 & 83.30          & 78.20          & 73.60          & 82.83         \\
\citet{seo2019learning}       & 95.87          & \textbf{84.67} & 77.65          & \textbf{82.23} & 85.11         \\
\citet{huang2020self}     & 95.99             & 83.43          & 80.31          & 80.85          & 85.15         \\
\rowcolor{Gray}
\textit{\textbf{This paper}} & 95.97 {\scriptsize $\pm$0.24} & 83.92 {\scriptsize $\pm$0.71} & \textbf{81.61} {\scriptsize $\pm$0.59} & 80.31 {\scriptsize $\pm$0.91} & \textbf{85.45} {\scriptsize $\pm$0.24} \\ 
\bottomrule
\end{tabular}
\vspace{-4mm}
\end{table*}

\textbf{Benefits of Synergistic Bayesian Invariant Learning.} 
The last three rows of Table~\ref{ablation} show the performance when introducing Bayesian \textit{and} invariant learning into both the classifier and feature extractor.
Both the Bayesian learning (row (h)) and deterministic invariant learning (row (i)) in two layers perform better than introducing the corresponding properties into only one layer of the model (compared with (b), (c) and (e), (f)) on most domains, as well as in terms of mean performance.
Overall, employing our Bayesian invariant layer in both the classifier and feature extractor achieves the best performance, as shown in row (j). 

The last row in Figure~\ref{fig:visual} further demonstrates the benefits of synergistic Bayesian invariant learning. 
Incorporating the Bayesian classifier and the Bayesian feature extractor further maximizes the inter-class distance of all domains, as comparing (h) to (b) and (e).
Introducing deterministic invariance into both the classifier and feature extractor also improves the domain-invariant property of features and the generalization to the target domain to some extent, as shown in subfigure (i).
Moreover, by utilizing both the Bayesian domain-invariant classifier and Bayesian domain-invariant feature extractor, our method combines their benefits shown in (d) and (g) and achieves the best performance, as shown in (j). 
This indicates the benefits of the synergy between Bayesian inference and probabilistic domain-invariant learning for domain generalization.

\vspace{2mm}
\textbf{Effect of Scale Mixture Priors.} 
Instead of using an uninformative diagonal Gaussian prior, we adopt the scale mixture prior \cite{Blundell2015WeightUI}. To show the effect of the prior, we conduct experiments with both the standard Gaussian prior and scale mixture Gaussian prior as in (\ref{scalemixture}).
The results are reported in Table~\ref{prior}.
The scale mixture prior achieves better performance on all domains.
We have also done an ablation on the scaling mixture prior by changing the value of $\pi$ in (\ref{scalemixture}). The prior with $\pi {=} 0.5$ achieves the best performance on the ``cartoon'' domain of PACS (Figure \textcolor{mydarkblue}{C.2} in supplementary), and thereby we use this value in all other experiments. 

\vspace{1mm}
\textbf{Effect of More Bayesian Layers.}
We also experiment with more Bayesian layers in the feature extractor, as shown in Table~\ref{2layers}.
The settings of the model in the first row are the same as row (j) in Table~\ref{ablation}.
When introducing another Bayesian layer $\bphi '$ without domain invariance into the model, as shown in the second row, the average performance improves slightly. 
However, if we introduce the Bayesian domain-invariant learning into $\bphi '$ (third row), the overall performance deteriorates slightly. 
This may due to the loss of information in the features caused by the excessive use of domain-invariant learning.
In addition, due to the Bayesian inference and Monte Carlo sampling, more Bayesian layers leads to higher memory usage and more computations (more detailed discussion in supplementary material).
As such, we prefer to apply the Bayesian invariant learning only in the last feature extraction layer and the classifier.

\begin{table*}[t]
\caption{Comparison on Office-Home. Our method achieves the best performance on the ``Art'' and ``Clipart'' domains, while being competitive on the ``Product'' and ``Real'' domains. Again we report the best overall mean accuracy.}
\vspace{1mm}
\centering
\label{office}
		\setlength\tabcolsep{4pt} 
\begin{tabular}{llllll}
\toprule
          & Art            & Clipart        & Product & Real           & \textit{Mean}       \\ \midrule
Baseline          &       54.84         &       49.85       &    72.40    &     73.14       &        62.55       \\
\citet{carlucci2019domain}            & 53.04          & 47.51          & 71.47   & 72.79          & 61.20          \\
\citet{li2018domain}  & 56.50          & 47.30          & 72.10   & 74.80          & 62.68          \\
\citet{seo2019learning}              & 59.37          & 45.70          & 71.84   & 74.68          & 62.90          \\
\citet{huang2020self}             & 58.42          & 47.90 & 71.63   & 74.54          & 63.12          \\
\citet{zhou2020learning}            & 60.60       & 50.10          & \textbf{74.80}   & \textbf{77.00}          & 65.63 \\
\rowcolor{Gray}
\textit{\textbf{This paper}}       & \textbf{61.81} {\scriptsize $\pm$0.36} & \textbf{53.27} {\scriptsize $\pm$0.37} & 74.27 {\scriptsize $\pm$0.35} & 76.31 {\scriptsize $\pm$0.24} & \textbf{66.42} {\scriptsize $\pm$0.18} \\   
\bottomrule
\end{tabular}
\vspace{-4mm}
\end{table*}

\begin{table*}[t]
\caption{Comparison on Rotated MNIST and Fashion-MNIST. In-distribution accuracy is evaluated on the test sets of MNIST and Fashion-MNIST with rotation angles of $15^\circ$, $30^\circ$, $45^\circ$, $60^\circ$ and $75^\circ$, while out-of-distribution accuracy is evaluated on test sets with angles of $0^\circ$ and $90^\circ$. Our method achieves best performance on both the in-distribution and out-of-distribution test sets.}
\vspace{2mm}
\centering
\label{mnist}
	\setlength\tabcolsep{4pt} 
\begin{tabular}
{lllll}
\toprule
& \multicolumn{2}{c}{\textbf{MNIST}} & \multicolumn{2}{c}{\textbf{Fashion-MNIST}} \\
\cmidrule(lr){2-3} \cmidrule(lr){4-5} 
 & In-distribution & Out-of-distribution  & In-distribution & Out-of-distribution\\ \midrule
Baseline   & 98.4  & 93.5 & 89.6     & 76.9    \\
\citet{dou2019domain}      & 98.2 & 93.2  & 86.9   & 72.4 \\
\citet{piratla2020efficient}     & 98.4 & 94.7 & 89.7  & 78.0      \\
\rowcolor{Gray}
\textit{\textbf{This paper}} & \textbf{99.0} {\scriptsize $\pm$0.02}    & \textbf{96.5} {\scriptsize $\pm$0.08}  & \textbf{91.5} {\scriptsize $\pm$0.10} & \textbf{83.5} {\scriptsize $\pm$0.63} \\ \bottomrule
\end{tabular}
\vspace{-4mm}
\end{table*}

\vspace{-2mm}
\paragraph{State-of-the-Art Comparisons.}
We compare our method with the state-of-the-art on four datasets. 
For comprehensive comparison, we also include a vanilla deep convolutional ResNet-18 network as a baseline, without any Bayesian treatment, as done in row (a) of Table~\ref{ablation}. The results for each dataset are reported in Tables~\ref{pacs}, \ref{office} and \ref{mnist}.

On PACS, in Table~\ref{pacs}, our method achieves the best mean accuracy. For each individual domain, we are competitive with the state-of-the-art and even exceed all other methods on the ``cartoon'' domain.
On Office-Home, in Table~\ref{office}, we again achieve the best mean accuracy. It is worth mentioning that on the most challenging ``art'' and ``clipart'' domains, we also deliver the highest accuracy, with a good improvement over previous methods. However, \citet{zhou2020learning} and \citet{seo2019learning} outperform the proposed model on some domains of PACS and Office-Home. 
In \cite{zhou2020learning}, the source domains are augmented by a generator that synthesizes data from pseudo-novel domains, which often have similar characteristics with the source data. 
This pays off when the target data also has similar characteristics to the source domains, as the pseudo domains are more likely to cover the target domain, as can be seen for ``product'' and ``real world'' in Office-Home and ``photo'' in PACS. 
When the test domain is different from all of the training domains the performance suffers,  e.g., ``clipart'' in Office-Home and ``sketch'' in PACS. 
We highlight that our method generates domain-invariant representations and classifiers, resulting in competitive results across all domains and overall.
In addition, \citet{seo2019learning} combine batch and instance normalization for domain generalization. This tactic is effective on PACS, but less so on Office-Home. 
We attribute this to the larger number of categories in Office-Home, where instance normalization is known to make features less discriminative with respect to object categories \citep{seo2019learning}.
In contrast, our Bayesian domain-invariant learning establishes domain-invariant features and predictions in a probabilistic form by introducing uncertainty into the model, resulting in good performance on both PACS and Office-Home.

On Rotated MNIST and Fashion-MNIST, following the experimental settings in \citet{piratla2020efficient}, we evaluate our method on the in-distribution and out-of-distribution sets. As shown in Table~\ref{mnist}, our method achieves the best performance on both sets of the two datasets. 
In particular, our method improves the classification performance on the out-of-distribution sets, demonstrating its strong generalizability to unseen domains, which is also consistent with the findings in Figure~\ref{fig:visual}.

\section{Conclusion}

In this work, we propose a variational Bayesian learning framework for domain generalization. 
We introduce Bayesian neural networks into the model,
which are able to better represent uncertainty and enhance the generalization to out-of-distribution data. To handle the domain shift between source and target domains, we propose a domain-invariant principle under the variational inference framework, which is incorporated by establishing a domain-invariant feature extractor and classifier. 
Our method combines the representational power of deep neural networks and uncertainty modeling ability of Bayesian learning, demonstrating effectiveness for domain generalization. Ablation studies further validate these benefits.
Our Bayesian invariant learning sets a new state-of-the-art on four domain generalization benchmarks.

\section*{Acknowledgements}
This work is financially supported by the Inception Institute of Artificial Intelligence, the University of Amsterdam and the allowance Top consortia for Knowledge and Innovation (TKIs) from the Netherlands Ministry of Economic Affairs and Climate Policy.
\nocite{langley00}

\bibliography{vil}
\bibliographystyle{icml2021}




\onecolumn








\appendix
\section{Derivation}
\label{sec:derive}

\subsection{Derivation of the Upper Bounds of Probabilistic Domain-invariant Learning}

To implement $\mathbb{E}_{q_{\zeta}} \big[ p_{\btheta}(\y_{\zeta}|\x_{\zeta}) \big] $
in a tractable way, we derive the upper bound in Section 2.2, which is achieved via Jensen's inequality:
\begin{equation}
\label{derive4}
\begin{aligned}
\dkl \big[ p_{\btheta}(\y_{i}|\x_{i})|| \mathbb{E}_{q_{\zeta}} [p_{\btheta}(\y_{\zeta}|\x_{\zeta})] \big]  & = \mathbb{E}_{p_{\btheta}(\y_{i}|\x_{i})} \big[\log p_{\btheta}(\y_i|\x_{i}) - \log \mathbb{E}_{q_{\zeta}} [p_{\btheta}(\y_{\zeta}|\x_{\zeta})] \big]\\ &
\leq \mathbb{E}_{p_{\btheta}(\y_i|\x_{i})} \big[\log p_{\btheta}(\y_i|\x_{i}) - \mathbb{E}_{q_{\zeta}} [\log p_{\btheta}(\y_{\zeta}|\x_{\zeta})]\big]\\ &
= \mathbb{E}_{q_{\zeta}} \Big[\dkl \big[ p_{\btheta}(\y_i|\x_{i})||p_{\btheta}(\y_{\zeta}|\x_{\zeta}) \big]\Big].
\end{aligned}
\end{equation}


\subsection{Derivation of Variational Bayesian Approximation for Feature Extractor ($\bphi$) and Classifier ($\psi$) layers}

We consider the model with two Bayesian layers $\bphi$ and $\bpsi$ as the last layer of the feature extractor and the classifier. The prior distribution of the model is $p(\bphi, \bpsi)$, and the true posterior distribution is  $p(\bphi, \bpsi|\x, \y)$.
Following the settings in Section 2.1, we need to learn a variational distribution $q(\bphi, \bpsi)$ to approximate the true posterior by minimizing the KL divergence from $q(\bphi, \bpsi)$ to $p(\bphi, \bpsi|\x, \y)$:
\begin{equation}
\begin{aligned}
    \bphi^{*}, \bpsi^{*} = \mathop{\arg\min}\limits_{\bphi, \bpsi} 
     \mathbb{D}_{\rm{KL}} \big[ q(\bphi, \bpsi)||p(\bphi, \bpsi|\x, \y)\big].
\end{aligned}
\end{equation}

By applying the Bayesian rule $p(\bphi, \bpsi|\x, \y) \propto p(\y|\x, \bphi, \bpsi)p(\bphi, \bpsi)$, the optimization is equivalent to minimizing:
\begin{equation}
\begin{aligned}
    &{\mathcal{L}}_{\rm{Bayes}} 
     = \int q(\bphi, \bpsi) \log \frac{q(\bphi, \bpsi)}{p(\bphi, \bpsi)p(\y|\x, \bphi, \bpsi)} \mathrm{d}\bphi \mathrm{d}\bpsi
     = \mathbb{D}_{\rm{KL}} \big[ q(\bphi, \bpsi)||p(\bphi, \bpsi)\big] - \mathbb{E}_{q(\bphi, \bpsi)} \big[\log p( \y|\x, \bphi, \bpsi) \big].
\end{aligned}
\end{equation}

With $ \bphi $ and $ \bpsi $ being independent,
\begin{equation}
\begin{aligned}
{\mathcal{L}}_{\rm{Bayes}}
&= 
-\mathbb{E}_{q(\bpsi)} \mathbb{E}_{q(\bphi)} [\log p(\y|\x, \bpsi, \bphi)] + \dkl[q(\bpsi)||p(\bpsi)] +
\dkl[q(\bphi)||p(\bphi)].
\end{aligned}
\end{equation}



\section{Details of Bayesian Domain-invariant Training}
\label{sec:detailtrain}

\begin{figure*}[t]
\centering
\includegraphics[width=.95\linewidth]{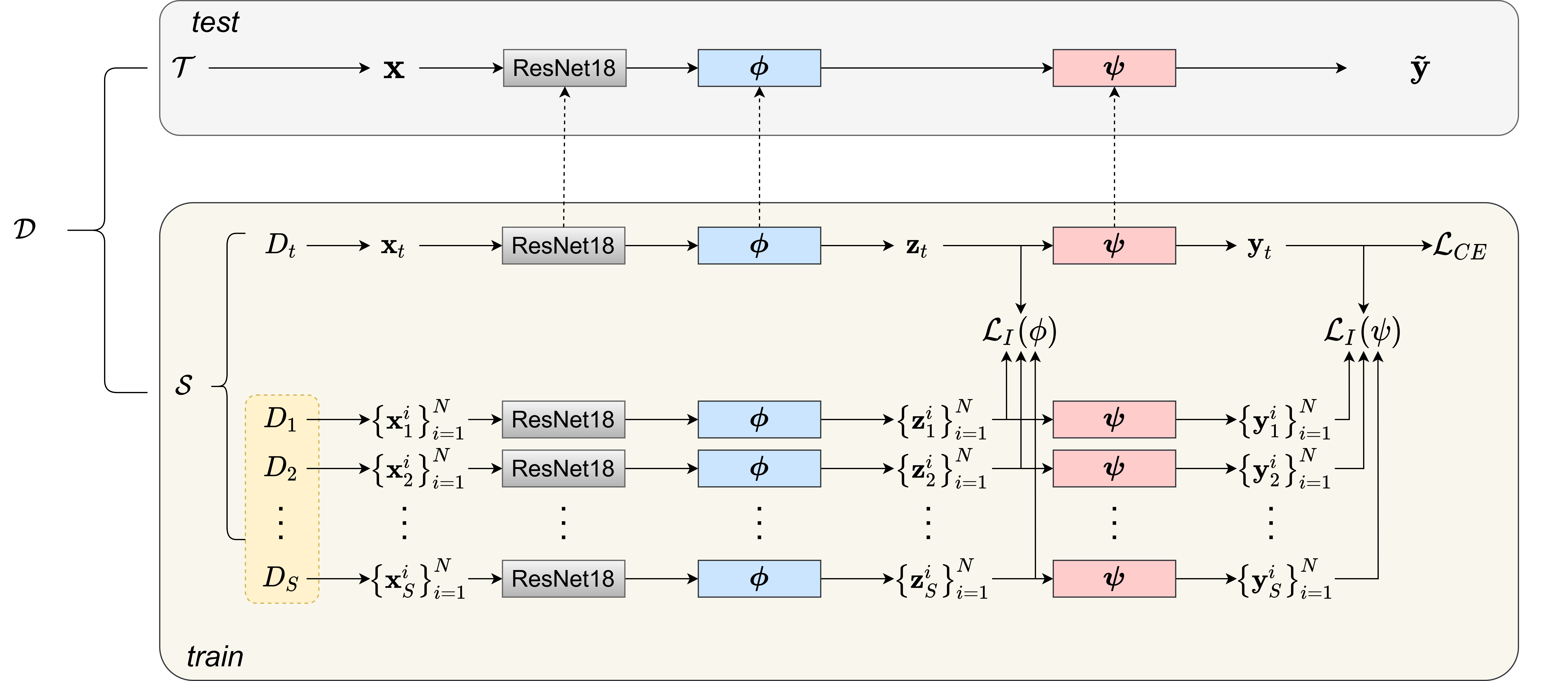}
\vspace{-2mm}
\caption{Illustration of the training phase of our Bayesian domain-invariant learning. 
$\mathcal{S}$ denotes the source domains, $\mathcal{T}$ denotes the target domains, and $\mathcal{D}=\mathcal{S}\cup\mathcal{T}$. 
$\x$, $\z$ and $\y$ denote inputs, features and outputs of samples in each domain.
$\mathcal{L}_I(\bpsi)$ and $\mathcal{L}_I(\bphi)$ denote the domain-invariant objective functions for the classifier and feature representations as detailed in (7) and (9) in the main paper.}

\label{fig:arch}
\vspace{-5mm}
\end{figure*}

In domain generalization, the given domains $\mathcal{D}{=}{\left\{ D_i \right\}}^{|\mathcal{D}|}_{i=1}$ are divided into the source domains $\mathcal{S}$ and the target domains $\mathcal{T}$.
During training, as the data from $\mathcal{T}$ is never seen, only the source domains $\mathcal{S}$ are accessible.
In each iteration, as shown in Figure~\ref{fig:arch}, we randomly choose one source domain from $\mathcal{S}$ as the meta-target domain $D_t$, and the rest of the source domains ${\left\{ D_s \right\}}^{S}_{s=1}$ are treated as the meta-source domains, where $S{=}|\mathcal{S}|-1$.
We randomly select a batch of samples $\x_t$ from $D_t$. 
We also select $N$ samples ${\left\{ \x^i_s \right\}}^{N}_{i=1}$, which are in the same category as $\x_t$, from each of the meta-source domains $D_s$. 
All these samples are sent to the Bayesian invariant feature extractor $\bphi$ after a ResNet18 backbone to get the representations $\z_t$ and ${\left\{\z^i_s \right\}}^{N}_{i=1}$, which are then sent to the Bayesian invariant classifier $\bpsi$ to get the predictions $\y_t$ and ${\left\{\y^i_s \right\}}^{N}_{i=1}$. 
We obtain the Bayesian invariant objective function for the feature extractor $\mathcal{L}_{I}(\bphi)$ by calculating the mean of the KL divergence of $p(\z_t|\x_t, \bphi)$ and each $p(\z^i_s|\x^i_s, \bphi)$ as (9) in the main paper. 
The Bayesian invariant objective function for feature classifier $\mathcal{L}_{I}(\bpsi)$ is calculated in a similar way on $p(\y_t|\z_t, \bpsi)$ and ${\left\{ p(\y^i_s|z^i_s, \bpsi) \right\}}^{N}_{i=1}$ as (7) in the main paper.
In addition to the Bayesian invariant objective functions, there are also a cross-entropy loss $\mathcal{L}_{CE}$ on $p(\y_t|\x_t, \bpsi, \bphi)$ and two Kullback-Leibler terms between the posteriors and priors of $\bpsi$ and $\bphi$ as detailed in (11) in the main paper. 

\section{Ablation Study for Hyperparamters}
\label{sec:param}

\begin{figure*}[t]
\centering
\includegraphics[width=\linewidth]{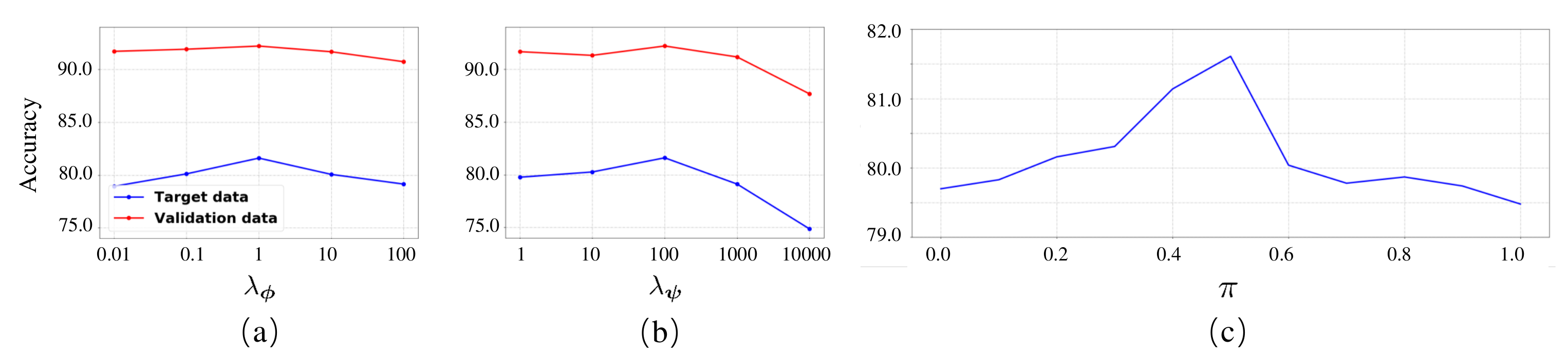}
\vspace{-6mm}
\caption{Performance on ``cartoon'' domain in PACS with different hyperparameters $\lambda_{\phi}$, $\lambda_{\psi}$ and $\pi$. The red line denotes the accuracy on validation data while the blue line denotes accuracy on target data. The optimal value of $\lambda_{\phi}$, $\lambda_{\psi}$ and $\pi$ are 1, 100 and 0.5.}
\label{fig:param}
\end{figure*}

We also ablate the hyperparameters $\lambda_{\bphi}$, $\lambda_{\bpsi}$ and $\pi$ to show their effects. The experiments are conducted on PACS by using \textit{cartoon} as the target domain. The results are shown in Figure~\ref{fig:param}. Specifically,
we produce Figure~\ref{fig:param} (a) by fixing $\lambda_{\bpsi}$ as 100 and adjusting $\lambda_{\bphi}$, Figure~\ref{fig:param} (b) by fixing $\lambda_{\bphi}$ as 1 and adjusting $\lambda_{\bpsi}$ and Figure~\ref{fig:param} (c) by adjusting $\pi$ while fixing other settings as in Section 4.1.
$\lambda_{\bphi}$ and $\lambda_{\bpsi}$ balance the influence of the Bayesian learning and domain-invariant learning, and their optimal values are 1 and 100, respectively. If the values are too small, the model tends to overfit to source domains as the performance on target data drops more obviously than on validation data.
By contrast, too large values of harm the overall performance of the model as there are obvious decrease of accuracy on both validation data and target data.
Moreover, $\pi$ balances the two components of the scale mixture prior of our Bayesian model. According to \cite{Blundell2015WeightUI}, the two components cause a prior density with heavier tail while many weights tightly concentrate around zero. Both of them are important. The performance is the best when $\pi$ is 0.5 according to Figure~\ref{fig:param} (c), which demonstrates the effectiveness of using two components in the scale mixture prior.

\section{Additional Ablations on PACS}
\label{extraexperiments}

To further demonstrate the effectiveness of the Bayesian inference and Bayesian invariant learning, we conduct some more supplementary experiments with other settings on PACS  as shown in Table~\ref{ablations}.
For more comprehensive comparisons, we add four more settings with IDs (k), (l), (m), and (n) to the settings in Table~\ref{ablation}.

Comparing (k) and (l) with (d), 
we find that when employing the Bayesian invariant classifier $\bpsi$ in the model, introducing Bayesian or deterministic invariant learning into the last layer of the feature extractor $\bphi$ both further improves the overall performance. 
Moreover, introducing Bayesian invariant learning into $\bphi$ achieves the best improvements as shown in row (j).
A similar phenomenon occurred on the classifier as comparing (m), (n) and (j) with (g).
When there is a Bayesian invariant layer in the last layer of the feature extractor, introducing Bayesian learning or deterministic invariant learning both improves the performance, while Bayesian invariant learning achieves the best result.
This further indicates the benefits of introducing uncertainty into the model and the invariant learning under the Bayesian framework.

\begin{table*}[t]
\caption{More detailed ablation study on PACS. 
Compared to Table~\ref{ablation} we add four more settings with IDs (k), (l), (m) and (n).
Bayesian inference benefits domain generalization.
Our Bayesian invariant learning achieves better performance than the deterministic ones. 
} 
\vspace{2mm}
\centering
\label{ablations}
		\setlength\tabcolsep{4pt} 
\begin{tabular}{cccccccccc}
\toprule
 & \multicolumn{2}{c}{\textbf{Classifier $\bpsi$}} & \multicolumn{2}{c}{\textbf{Feature extractor $\bphi$}} & \multicolumn{5}{c}{\textbf{PACS}}\\
\cmidrule(lr){2-3} \cmidrule(lr){4-5} \cmidrule(lr){6-10}
ID &
  Bayesian &
  Invariant &
  Bayesian &
  Invariant &
  Photo &
  Art-painting &
  Cartoon &
  Sketch &
  \textit{Mean} \\ 
  \midrule
(a) & $\times$     & $\times$     & $\times$     & $\times$  & 92.85 {\scriptsize $\pm$0.21}   & 75.12 {\scriptsize $\pm$0.48} & 77.44  {\scriptsize $\pm$0.26} & 75.72 {\scriptsize $\pm$0.47} & 80.28 {\scriptsize $\pm$0.42} \\
\hline
(b) & $\checkmark$ & $\times$     & $\times$     & $\times$  & 93.89 {\scriptsize $\pm$0.29} & 77.88 {\scriptsize $\pm$0.53} & 78.20 {\scriptsize $\pm$0.39} & 77.75 {\scriptsize $\pm$0.75} & 81.93 {\scriptsize $\pm$0.22}\\
 (c) &
  $\times$ &
  $\checkmark$ &
  $\times$ &
  $\times$ & 93.95 {\scriptsize $\pm$0.51} & 80.03 {\scriptsize $\pm$0.72}  & 78.03 {\scriptsize $\pm$0.77} & 77.83 {\scriptsize $\pm$0.52} & 82.46 {\scriptsize $\pm$0.67}
   \\
 (d) &
  $\checkmark$ &
  $\checkmark$ &
  $\times$ &
  $\times$ &
  95.21 {\scriptsize $\pm$0.26}  & 81.25 {\scriptsize $\pm$0.76} & 80.67 {\scriptsize $\pm$0.73} & 79.31 {\scriptsize $\pm$0.94} & 84.11 {\scriptsize $\pm$0.39} \\ 
  \hline
(e) & $\times$     & $\times$ & $\checkmark$ & $\times$ & 92.81 {\scriptsize $\pm$0.35} &
  78.66 {\scriptsize $\pm$0.56} &
  77.90 {\scriptsize $\pm$0.40} &
  78.72 {\scriptsize $\pm$0.86} &
  82.02 {\scriptsize $\pm$0.26}  \\
 (f) & $\times$     & $\times$ & $\times$     & $\checkmark$   & 94.17 {\scriptsize $\pm$0.35} & 79.75 {\scriptsize $\pm$0.68}  & 79.51 {\scriptsize $\pm$0.98} & 78.31 {\scriptsize $\pm$1.11} & 82.94 {\scriptsize $\pm$0.53} \\
 (g) & $\times$     & $\times$ & $\checkmark$     & $\checkmark$ & 95.15 {\scriptsize $\pm$0.26} & 80.96 {\scriptsize $\pm$0.69} & 79.57 {\scriptsize $\pm$0.85} & 79.15 {\scriptsize $\pm$0.98} & 83.71 {\scriptsize $\pm$0.65}\\ \hline
 (m) & $\checkmark$     & $\times$ & $\checkmark$     & $\checkmark$ & 95.87 {\scriptsize $\pm$0.48} & 81.15 {\scriptsize $\pm$0.49} & 79.39 {\scriptsize $\pm$0.33} & 80.15 {\scriptsize $\pm$1.16} & 84.14
{\scriptsize $\pm$0.41} \\
 (n) & $\times$     & $\checkmark$ & $\checkmark$     & $\checkmark$ & 95.39 {\scriptsize $\pm$0.12} & 82.32 {\scriptsize $\pm$0.37} & 80.27 {\scriptsize $\pm$0.58} & 79.61 {\scriptsize $\pm$0.93} & 84.40 {\scriptsize $\pm$0.45}
 \\ \hline
(h) & $\checkmark$ & $\times$ & $\checkmark$     & $\times$  & 93.83 {\scriptsize $\pm$0.19} & 82.13 {\scriptsize $\pm$0.41} & 79.18 {\scriptsize $\pm$0.48} & 79.03 {\scriptsize $\pm$0.78} & 83.54 {\scriptsize $\pm$0.34} \\
 (i) & $\times$     & $\checkmark$     & $\times$ & $\checkmark$ & 94.12 {\scriptsize $\pm$0.22} & 80.52 {\scriptsize $\pm$0.61} & 80.39 {\scriptsize $\pm$0.81} & 78.53 {\scriptsize $\pm$0.95} & 83.39 {\scriptsize $\pm$0.52} \\
 (j) &
  $\checkmark$ &
  $\checkmark$ &
  $\checkmark$ &
  $\checkmark$ &
  \textbf{95.97} {\scriptsize $\pm$0.24}&
  \textbf{83.92} {\scriptsize $\pm$0.71}&
  \textbf{81.61} {\scriptsize $\pm$0.59}&
  \textbf{80.31} {\scriptsize $\pm$0.91}&
  \textbf{85.45} {\scriptsize $\pm$0.24}\\ 
  \bottomrule
\end{tabular}
\end{table*}
\vspace{-2mm}

\section{Extra Visualizations}
\label{sec:visual}

\begin{figure*}[t]
\centering
\includegraphics[width=\linewidth]{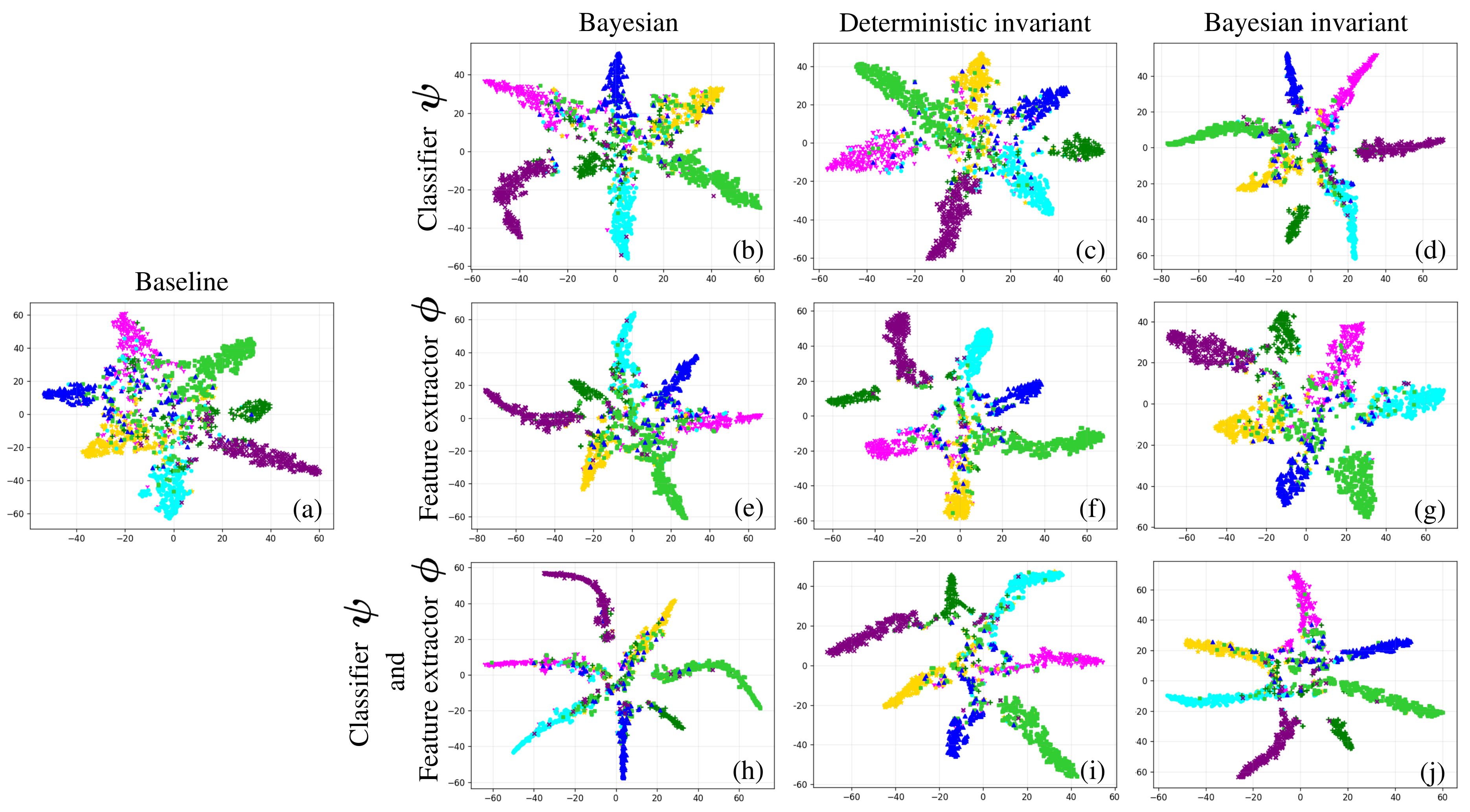}
\caption{Visualization of feature representations of the target domain. The subfigures have the same experimental settings as the experiments in Table~\ref{ablation} and Figure~\ref{fig:visual} in the main paper. To obtain more obvious contrast between the subfigures, we use different colors to denote different categories.
The target domain is ``art-painting'', the same as in Figure~\ref{fig:visual}. We obtain a similar conclusion to Section 4.2, where by introducing uncertainty into both the model and the domain-invariant learning, our Bayesian invariant learning achieves better performance and generalization to the target domain.}
\label{fig:target}
\end{figure*}

\begin{figure*}[t]
\centering
\includegraphics[width=\linewidth]{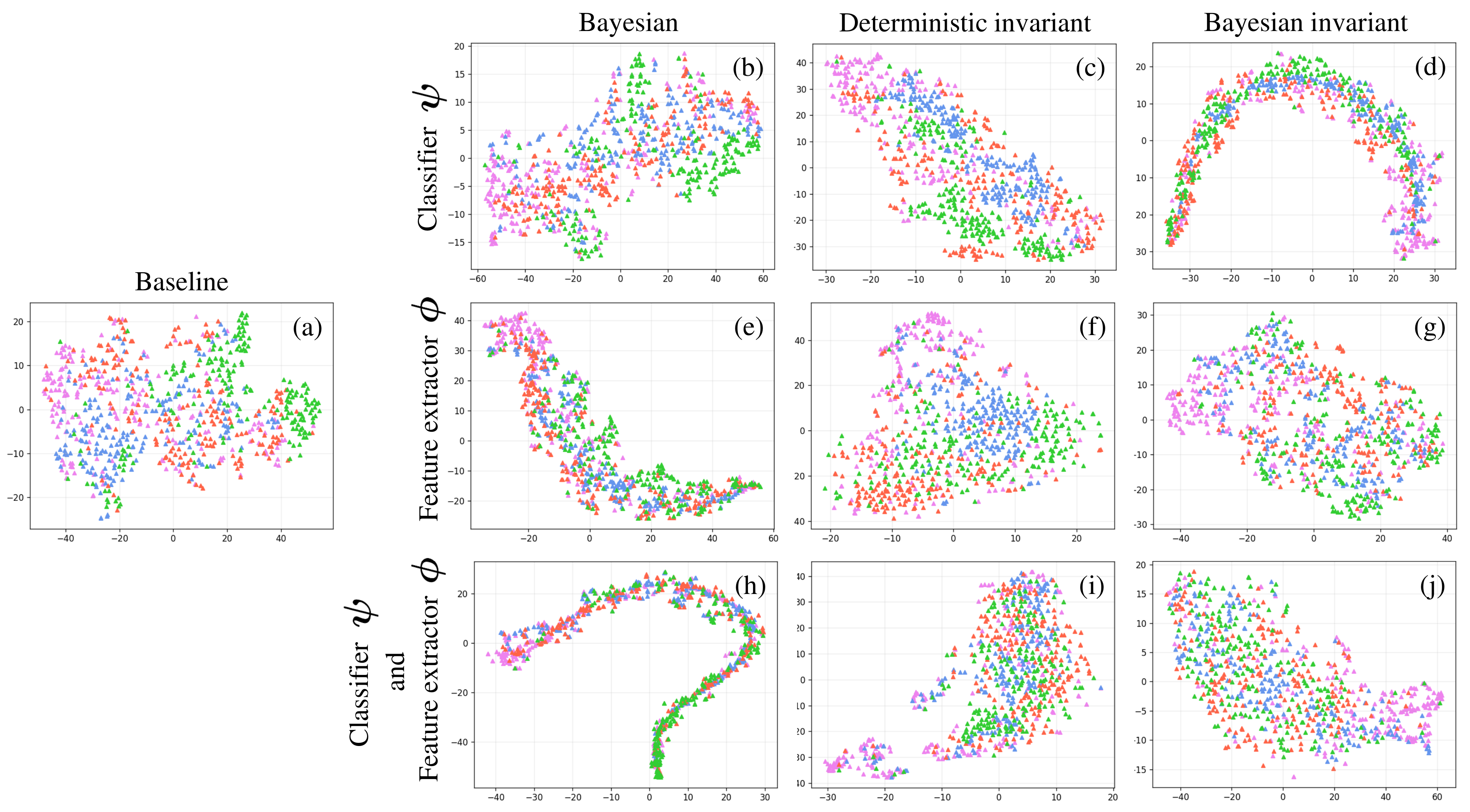}
\caption{Visualization of feature representations of one category. All samples are from the ``horse'' category with colors denoting different domains. The target domain is ``art-painting'' (violet). 
The ``Bayesian'' column shows Bayesian inference benefits domain generalization by gathering features from different domains to the same manifold. 
Comparing the subfigures in the ``Bayesian invariant'' column to those in the ``Deterministic invariant'' column indicates that our Bayesian invariant learning has better generalization to the target domain than the deterministic one.
}
\label{fig:horse}
\end{figure*}

To further observe and analyze the benefits of the Bayesian invariant learning, we conduct more detailed visualizations of the features in Figure~\ref{fig:target} and Figure~\ref{fig:horse}.
Visualizations in Figure~\ref{fig:target} show the features of all categories from the target domain only, while visualizations in Figure~\ref{fig:horse} show features of only one category from all domains. 
Similar to Figure~\ref{fig:visual} in Section 4.2 of the main paper, the visualization is conducted on the PACS dataset and the target domain is ``art-painting''. The chosen category in Figure~\ref{fig:horse} is ``horse''. 
We also visualize the features on the rotated MNIST and Fashion MNIST datasets in Figure~\ref{fig:mnist}. 
The results further demonstrate the effectiveness of our Bayesian invariant learning on both in-distribution data and out-of-distribution data.

\subsection{Visualizations on the Target Domain of PACS}
Figure.~\ref{fig:target} provides a more intuitive observation of the benefits of Bayesian domain-invariant learning on the target domain. 
\begin{itemize}
\item 
\textit{Bayesian learning benefits classification on the target domain as shown in the ``Bayesian'' column of subfigures.}
Comparing (b), (c) to (a), it is obvious that by introducing uncertainty into the model, both the Bayesian learning on the classifier and feature extractor enlarge the inter-class distance on the target domain. 
Samples from different categories tend to align on a thinner line, which makes them easier to classify.
This phenomenon is more obvious when introducing Bayesian invariant learning into both the classifier and the feature extractor, as shown in (h).
\item
\textit{Bayesian invariant learning is better than deterministic invariant learning.} 
Comparing the subfigures in the ``Bayesian invariant'' column to those in the ``Deterministic invariant'' column demonstrates the benefits of our Bayesian invariant learning. 
The Bayesian invariant classifier achieves larger inter-class distances than the deterministic invariant one, as shown in (d) and (c).
The Bayesian invariant feature extractor pushes the features in all categories away from the center as shown in (g), which also results in larger inter-class distance.
Moreover, employing the Bayesian invariant learning to both the classifier and the feature extractor conducts more obvious improvements, as comparing (j) to (i).
\item
\textit{Bayesian invariant learning benefits generalization to the target domain by introducing uncertainty into both model and domain-invariant learning.}
It is interesting to look at subfigures (b), (d), and (h). 
The visualizations in (d) are similar to (h), which also occurs in Figure~\ref{fig:visual} in the main paper.
Compared to (b), (h) introduces more uncertainty by employing Bayesian inference in the feature extractor, while (d) applies the Bayesian invariant learning based on the Bayesian classifier.
The similar observation in comparing (d) to (h) indicates that the benefits of our Bayesian invariant learning can be attributed to the uncertainty introduced into the domain-invariant learning.
The objective function in (6) in the main paper also indicates this conclusion.
Note that although both introduce uncertainty into the domain-invariant learning,
visualization in (g) is different from (d).
This is due to the uncertainty in the domain-invariant classifier and domain-invariant features having different effects on the feature space.
As shown in (d) and (g), both the Bayesian invariant classifier and feature extractor enlarge the inter-class distance on the target domain, though from different directions.
\end{itemize}

\subsection{Visualizations of Samples in One Class from Different Domains on PACS}
Figure~\ref{fig:horse} provides a deeper insight into the intra-class feature distributions of the same category from different domains. 
\begin{itemize}
\item 
\textit{Introducing uncertainty into the model gathers features from different domains to the same manifold.}
By comparing (b), (e), and (h) to (a), we find that introducing Bayesian inference into the model tends to gather the features to a certain manifold. This is even more obvious with more Bayesian layers as shown in (h).
The same phenomenon can be found in subfigure (d), where uncertainty is introduced in both the classifier and the domain-invariant learning.
Gathering the features into a certain manifold is beneficial to domain generalization as it facilitates classification, both on the source domains and the target domain.
\item
\textit{Bayesian invariant learning enables better generalization to the target domain than the deterministic counterpart.}
The deterministic invariant learning minimizes the distance between deterministic samples from different source domains.
That tends to be overfitting on the source domains. 
As shown in (c), (f), and (i), the samples from source domains are clustered well, while some samples from the target domains tend to be out of the cluster. There is even an obvious gap when applying deterministic invariant learning on both classifier and feature extractor, as shown in (i).
By contrast, our Bayesian invariant learning introduces uncertainty into domain-invariant learning. 
Thus, it achieves better mixture of features from the source domains and the target domain as shown in (d), (g), and (j). 
This indicates the Bayesian invariant learning enables better generalization to the target domain than the deterministic one.
\end{itemize}

\subsection{Visualizations on Rotated MNIST and Fashion-MNIST}

\begin{figure*}[t]
\centering
\includegraphics[width=.9\linewidth]{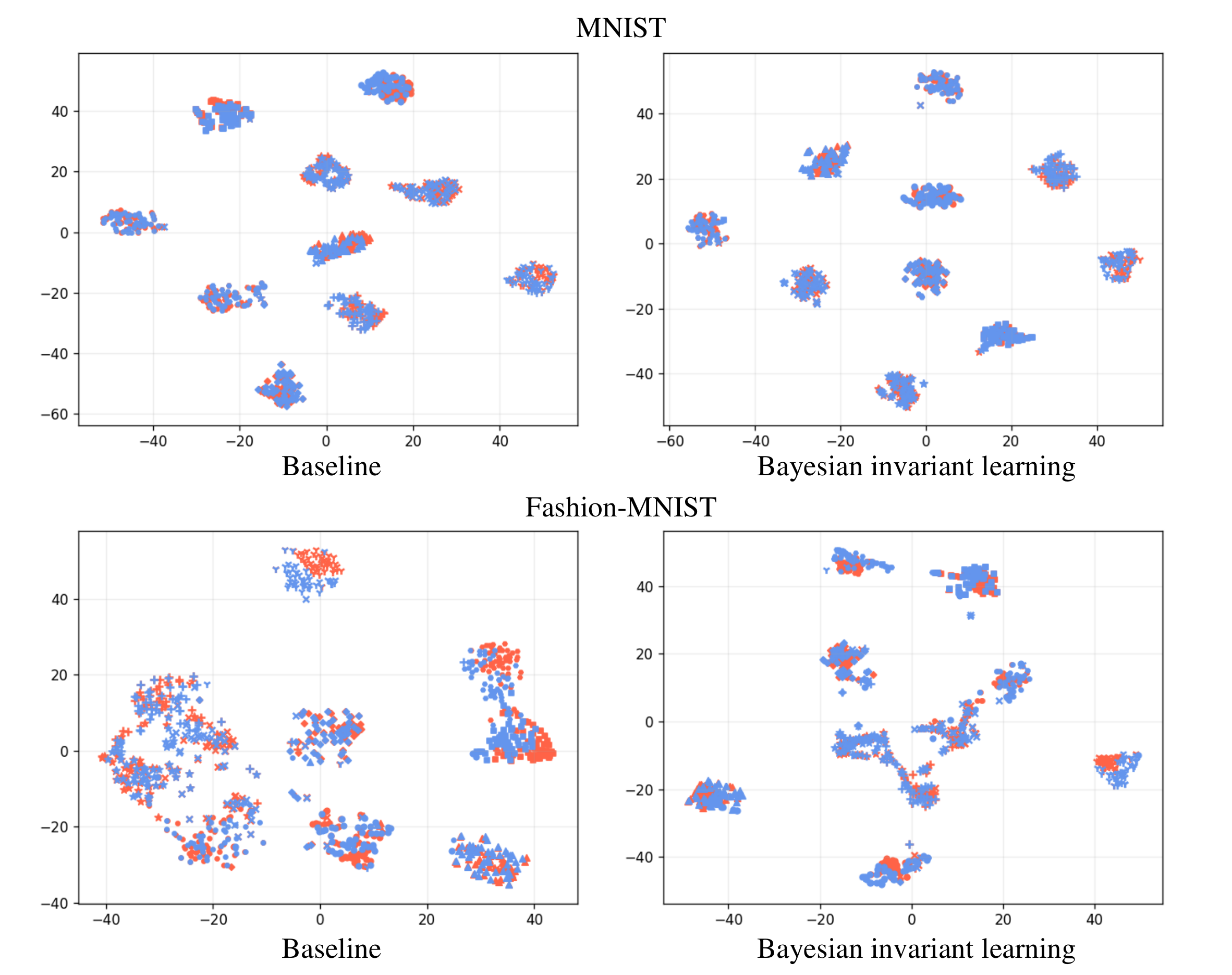}
\vspace{-3mm}
\caption{Visualization of feature representations in rotated MNIST and rotated Fashion-MNIST datasets. Samples from the in-distribution and out-of-distribution sets are in red and blue, respectively. Different shapes denote different categories. Compared to the baseline, our Bayesian invariant learning achieves better performance on both the in-distribution and out-of-distribution sets in each dataset, and especially on the out-of-distribution set from the Fashion-MNIST benchmark.}
\label{fig:mnist}

\end{figure*}

To further demonstrate the effectiveness of our Bayesian invariant learning, we also visualize the features on rotated MNIST and Fashion MNIST, as shown in Figure~\ref{fig:mnist}. 
Different shapes denote different categories. Red samples denote features from the in-distribution set and blue samples denote features from the out-of-distribution set. 
Compared with the baseline, our method reduces the intra-class distance between samples from the in-distribution set as well as the out-of-distribution set, and clusters the out-of-distribution samples of the same categories better, especially in the rotated Fashion-MNIST dataset.

\section{Computational Cost Analysis of Different Number of Bayesian Layers}

\begin{table}[!]
\caption{Computational cost in FLOPs, parameters and GPU memory usage for different number of Bayesian layers in the model.}
\vspace{1mm}
\centering
\label{flops}
{%
		\setlength\tabcolsep{4pt} 
\begin{tabular}{crrrr}
\toprule
Number of Bayesian layers & Monte Carlo samples & Extra FLOPs (M) & Extra parameters (M) & Memory usage (G)\\
\midrule
0 & 0 & 0 & 0 & 19.4\\ 
\hline
1 & 10 & 0.04 & 0.004 & 19.4\\
2 & 10 & 2.98 & 0.530 & 19.5\\
3 & 10 & 32.42 & 1.060 & 22.6\\
\hline
1 & 15 & 0.05 & 0.004 & 19.4 \\
2 & 15 & 4.74 & 0.530 & 19.8 \\
3 & 15 & 75.01 & 1.060 & 25.5 \\
\hline
1 & 20 & 0.07 & 0.004 & 19.4 \\
2 & 20 & 6.68 & 0.530 & 20.4 \\
3 & 20 & 138.77 & 1.060 & $>$ 31.7 \\

\bottomrule
\end{tabular}
}
\vspace{-2mm}
\end{table}

To quantify the effect of the Bayesian framework on the computations and parameters of the model, we show the FLOPs and parameters for different numbers of Bayesian layers in Table~\ref{flops}. 
The parameters of Bayesian layers are irrelevant to the batch size and the number of Monte Carlo samples. 
As the number of Bayesian layers increases, the extra parameters brought by the Bayesian layers grow  smoothly, which will not cause much trouble.
However, the extra FLOPs will have a significant increase with more Bayesian layers in the model.
In addition, the FLOPs are related to the number of Monte Carlo samples and batch size. As shown in the third column, a bigger number of Monte Carlo samples leads to much higher FLOPs, especially with more Bayesian layers in the model.
The extra FLOPs in the table are based on one image with size (224, 224, 3). When the batchsize becomes bigger, the FLOPs will also increase.

To be more intuitive, we also show the GPU memory usage of different numbers of Bayesian layers. 
The experiments are conducted on the PACS dataset based on a single Tesla V100 GPU. The batch size is 128, and the number of samples per category and meta-source domain is 32. 
The GPU memory increases slightly when employing one or two Bayesian layers into the model, especially when the number of Monte Carlo samples is small.
However, when introducing three Bayesian layers into the model, 
there will be a significant increase in GPU memory usage.
When the number of Monte Carlo samples is bigger, such as 20, the model with three Bayesian layers will even run out of the GPU memory during training, as shown in the last row.

Based on Table~\ref{flops}, we finally apply the Bayesian invariant learning only in the last feature extraction layer and the classifier. The batch size in our experiment is set to 128 and the number of Monte-Carlo sampling in each Bayesian layer is 10.

\section{Failure Cases}

\begin{figure*}[t]
\centering
\includegraphics[width=.9\linewidth]{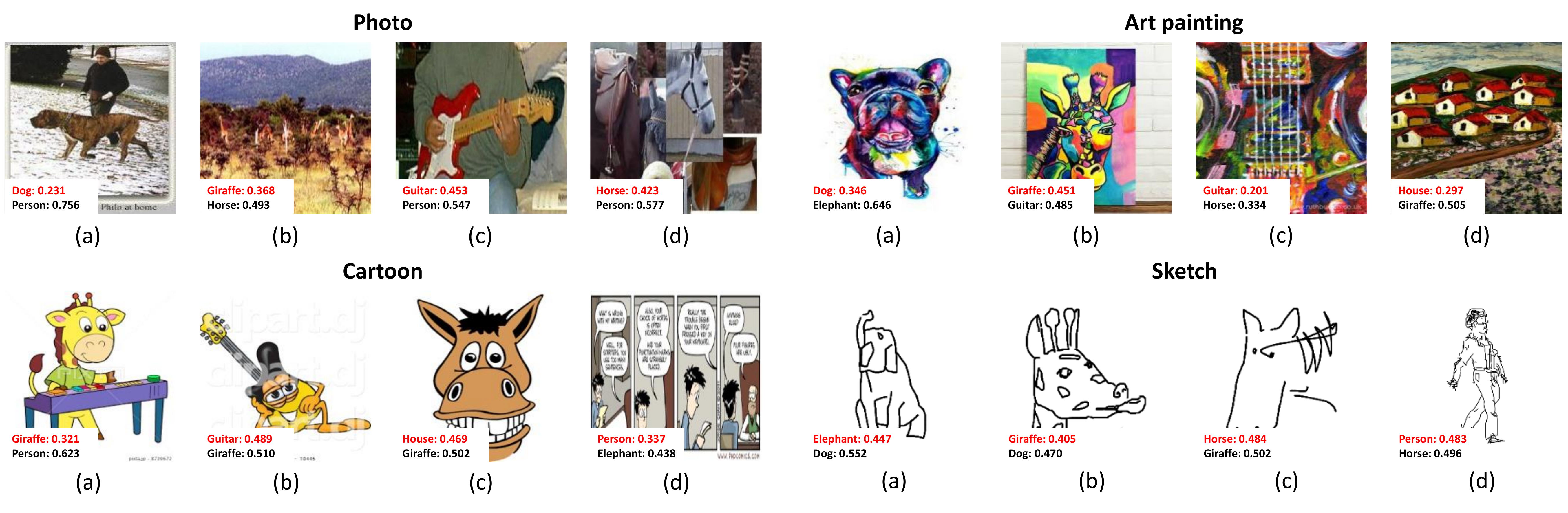}
\vspace{-3mm}
\caption{Some failure cases of our Bayesian invariant learning on PACS.
The numbers associated with each image are the top two prediction probabilities of our method, with ground truth labels in
red.
Our method fails to make the correct predictions, but provides high probabilities for the true label, indicating the effectiveness of introducing uncertainty.
}
\label{fail}
\end{figure*}

Finally, we consider several failure cases of our method. As shown in Figure~\ref{fail},
a complex scene makes it challenging to make correct predictions, e.g., (b), (d) in the ``photo'' domain and (d) in the ``cartoon'' domain. The failure cases in the ``art painting'' domain show the generalization performance of the method still needs further improvements, especially for samples with a very special and specific domain appearance. Our method also gets confused when samples have objects of different categories in the same image, e.g., (a) and (c) in the ``photo'' domain, or cluster the characteristics of different categories to the same sample, e.g., (a) and (b) in the ``cartoon'' domain and ``sketch'' domain. Although our Bayesian invariant learning fails to make the correct predictions in these challenging cases, it does provide reasonable probabilities for possible categories, which we attribute to the introduced model uncertainty and domain-invariant learning.

\nocite{langley00}





\end{document}